%% file: ICIP_2026.tex
\documentclass{article}
\usepackage{spconf}
\usepackage{graphicx,amsmath,amsfonts,amssymb,subcaption,cite}
\usepackage{hyperref}
\usepackage{booktabs}
\usepackage{xcolor}
\usepackage{multirow}

\usepackage{tikz}

\newcommand\copyrighttext{%
  \footnotesize \textcopyright 2026 IEEE. Personal use of this material is permitted.  Permission from IEEE must be obtained for all other uses, in any current or future media, including reprinting/republishing this material for advertising or promotional purposes, creating new collective works, for resale or redistribution to servers or lists, or reuse of any copyrighted component of this work in other works.}
\newcommand\copyrightnoticee{%
\begin{tikzpicture}[remember picture,overlay]
\node[anchor=south,yshift=10pt] at (current page.south) 
  {\fbox{\parbox{\dimexpr\textwidth-\fboxsep-\fboxrule\relax}{\copyrighttext}}};
\end{tikzpicture}%
}

\title{Authentication of Copy Detection Patterns via Cross-Camera Dual-Synthetic Referencing}

\name{}

\address{}

\name{Ivan Oleksiyuk, Roman Chaban, Slava Voloshynovskiy}
\address{Department of Computer Science, University of Geneva, Switzerland \\
\{ivan.oleksiyuk, roman.chaban, svolos\}@unige.ch}

\begin{document}

\topmargin=0mm

\maketitle 
\begin{abstract}
\input{sections/abstract}
\copyrightnoticee
\end{abstract}
\begin{keywords}
Copy Detection Patterns, authentication, synthetic referencing, mutual information, copy attacks
\end{keywords}

\input{sections/Section-introduction}

\input{sections/Sections-data}

\input{sections/Sections-methodology}

\input{sections/Section-IT}

\input{sections/Secrion-results}

\input{sections/Section-conclusion}

\newpage

\bibliographystyle{IEEEbib}
\bibliography{bibliography.bib}

\newpage

\appendix

\input{sections/Section-multi-camera}

\end{document}

%% file: sections/abstract.tex
Copy Detection Patterns (CDPs) are structures printed on physical objects to enable cost-effective authentication. Verification is achieved by comparing a captured image with the digital template from which the CDP was printed. In practice, printer stochasticity and camera distortions hinder this comparison, limiting robustness against counterfeiting. Prior work addressed camera effects by synthesising reference images in the verification camera domain, but it ignored printing variability. We introduce an enrolment-based cross-camera dual-synthetic referencing framework. Each printed CDP is first captured by a controlled enrolment camera, and a deep-learning-based translator jointly exploits the digital template and the enrolled capture to generate a high-quality reference for the verification image. We provide an information-theoretic justification showing that the dual reference is more informative than template-based references. Experiments on heterogeneous mobile cameras demonstrate improved authentication performance, robustness to machine-learning-based copy attacks, and reliable verification from small CDP regions and on low-end devices.

%% file: sections/Section-introduction.tex
\section{Introduction}
\label{sec:intro}

Copy Detection Patterns (CDPs) are binary structures printed on objects for authentication against counterfeiting. Their strength lies in combining low cost, scalability, and strong resistance to cloning \cite{picard2004,picard2008copy}. In recent years, several methods have sought to further enhance CDP counterfeiting detection through the use of deep-learning techniques \cite{chaban2021fakes, SyukronPratama_2023, computers12090183, app13148101, 10374744, chaban2024wifs, 11149957}.

\begin{figure}[tb]
    \centering
    \includegraphics[width=0.95\linewidth]{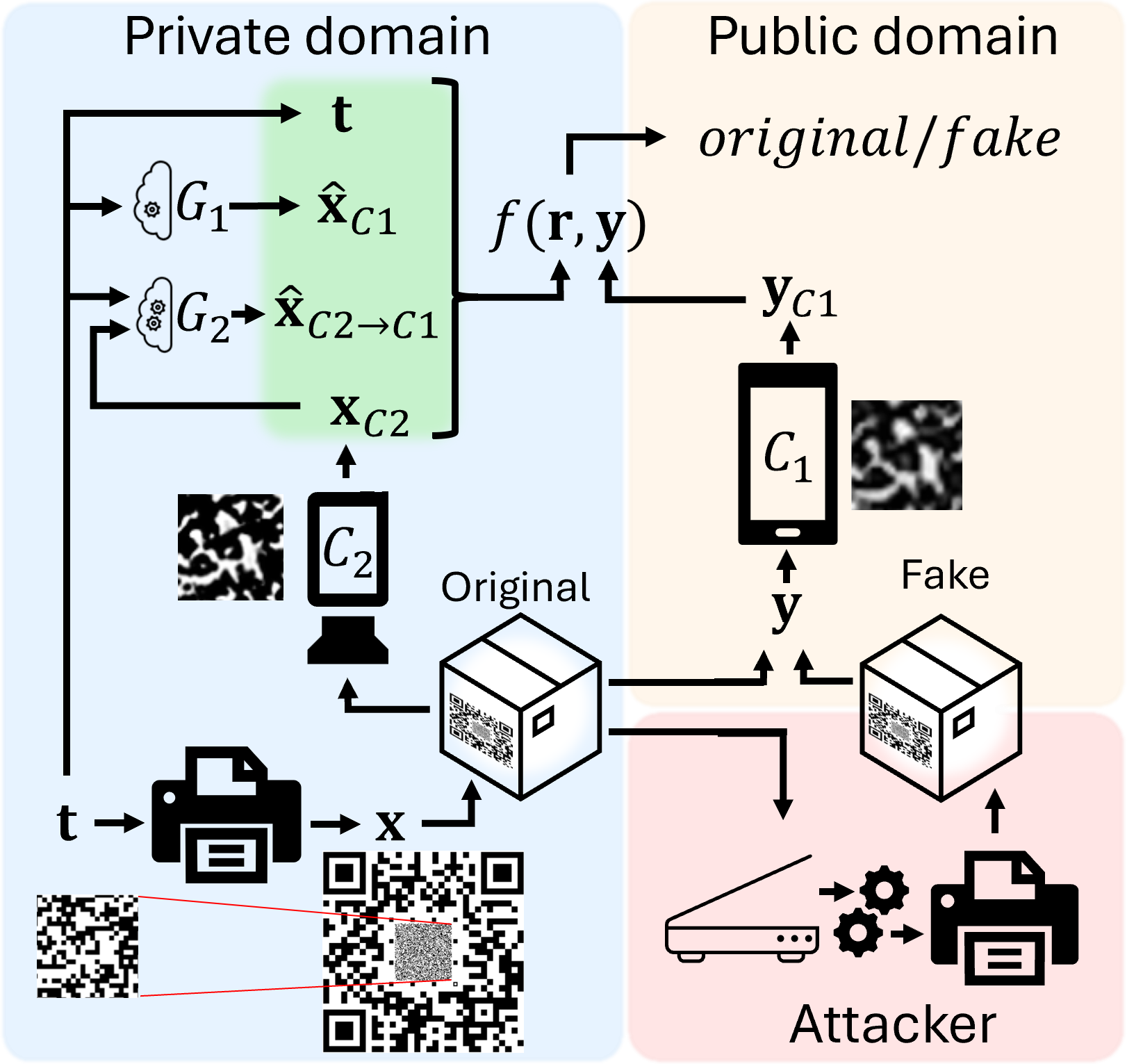}
    \caption{Diagram demonstrating the lifecycle of a CDP $\mathbf{x}$ integrated into a QR code as suggested by Scantrust \cite{scantrust}, including its production from digital template $\mathbf{t}$, enrolment with camera $C_2$, potential attack and verification using camera $C_1$.}
    \label{fig:scheme}
\end{figure}

The lifecycle of a CDP and different methods of its reference-based authentication are shown in Fig. \ref{fig:scheme}. First, a random digital template $\mathbf{t}$ is generated and used to print a physical CDP ${\bf x}$. The digital template is stored and serves as the reference during verification. In practice, a verifier captures an image ${\bf y}_{C_1}$ of the printed CDP with a mobile camera $C_1$ and compares it to $\mathbf{t}$ in some metric space to produce a decision on the authenticity of the CDP and object. However, during reproduction of $\bf t$ on the object surface, printers introduce stochastic dot gain (random ink spreading), while cameras add blur, noise, and compression. These degradations, absent in $\mathbf{t}$, make direct comparison unreliable, especially for printers with high dot gain and low-end devices, where imaging distortions are severe.

This weakness is amplified by recent machine learning attacks \cite{chaban2021fakes}, which can recover the digital template $\mathbf{t}$ from a single high-quality scan and produce high-quality copies. 

To address this, prior work \cite{chaban2024wifs} proposed transforming $\mathbf{t}$ into a synthetic reference $\hat{\mathbf{x}}_{C_1}$ that simulates a CDP as seen by camera $C_1$. While this approach models imaging distortions, it neglects printer stochasticity, so authentication performance remains limited, especially for small pattern sizes or low-resolution devices.

In this paper, we extend the referencing paradigm with an enrolment stage. At production, each printed CDP is captured by a controlled enrolment camera $C_2$, yielding an image ${\bf x}_{C_2}$ that naturally encodes some of the printer’s stochastic signature absent in the digital template. We then introduce a dual-synthetic cross-camera translator that fuses $({\bf t}, {\bf x}_{C_2})$ to generate a reference $\hat{{\bf x}}_{C_2 \rightarrow C_1}$ in the domain of the verification camera $C_1$. This dual-referencing strategy bridges the gap between digital design and physical realisation.

Our goal is to demonstrate that incorporating both the digital template $\bf t$ and the enrolled capture ${\bf x}_{C_2}$, which might be significantly different in statistics to ${\bf x}_{C_1}$,  yields a reference that is strictly more informative about the verification capture ${\bf x}_{C_1}$ than $\bf t$ alone. This claim is formalised using mutual information arguments and variance reduction inequalities, showing that the mutual information between ${\bf x}_{C_1}$ and $({\bf t}, {\bf x}_{C_2})$ is greater or equal to those between ${\bf x}_{C_1}$ and $\bf t$ with equality only when ${\bf x}_{C_2}$ provides no addition information about printer variability. 

Empirical results confirm this theoretical insight: the cross-camera approach markedly improves authentication accuracy on both low- and high-end mobile devices, withstands machine learning attacks, and remains effective even for small-size CDPs. By jointly exploiting digital and physical references, it enhances the robustness of CDP-based authentication.

%% file: sections/Sections-data.tex
\section{Dataset}
\label{sec:data}

In this study, we use an open dataset described in details in \cite{chaban2024wifs} \footnote{Dataset is available at \url{https://github.com/romaroman/cdp-synthetics-dataset}.}. The dataset consists of randomly generated digital templates of size $228\times228$ with equal probability of each cell being either black or white. CDPs are printed using an industrial-grade digital offset printer \textbf{HP Indigo 5500}. 

Then the CDPs are scanned using different mobile phone cameras. We select the wide-angle camera of iPhone XS as the lowest quality imaging device and the macro camera of iPhone 15 Pro Max as the highest quality camera in the dataset, as is identified in \cite{chaban2024wifs}. 

To produce ML-grade copies, for each CDP, an estimated digital template was obtained through methodology described in \cite{chaban2021fakes} which involved using a domain-specific pre-trained U-Net. Following the original methodology, the digital estimates were printed again using same set of conditions and parameters on the same industrial printer \textbf{HP Indigo 5500}. This process yielded a subset of fake $\mathbf{y}^{fake}_{C1}$ CDPs which are a counterpart to our main subject of the study in this scope, i.e., genuine CDPs $\mathbf{y}^{original}_{C1}$.

Conclusively, each existing physical (both original and fake) CDP was scanned multiple times with each mobile device. To assess the performance of a reference $\mathbf{r}$ taken by the same camera as the validation camera $C_{1}$, we use the first capture as probe $\mathbf{y}_{C1}$ and the second capture (only for original CDPs) as an enrolled reference $\mathbf{x}_{C1}$. In other scenarios we only use the first capture for each CDP.

%% file: sections/Sections-methodology.tex
\section{Methodology}
\label{sec:method}
\subsection{General setup}

All images are upscaled or downscaled to 3 times the digital template size, depending on the mobile device, namely, $684\times684$ pixels. 

For the training of all networks used in the scope of this work, we use overlapping patches of $128\times128$ pixels cropped from each training image with a vertical and horizontal stride of 64. Each patch is normalised to an amplitude range $[0, 1]$. For the training process, we have utilised a common set of affine non-destructive transformations to increase the number of samples, such as flips and rotations, which were applied to pairs of samples simultaneously.

\subsection{Synthetic CDP generator}
\label{subsec:synt}
To ensure direct comparison with previous methods, we use the model provided in \cite{chaban2024wifs} to obtain the synthetic CDPs $\mathbf{\hat{x}}_{C1,L_{comb}}=G_{C_1}(\mathbf{t})$ by training the generator $G_{C_1}$ for each camera. 
Although the loss $L_{comb}$ proposed in \cite{chaban2024wifs} consists of multiple loss components with different weighting factors, our empirical analysis indicates that the $L_2$ term is the dominant contributor to the overall performance. 
Therefore, as an ablation study, we reimplemented the synthetic reference with a conditional generator $\mathbf{\hat{x}}_{C1, L_2}=G_1(\mathbf{t}, C_1)$ that only minimises the $L_2$ loss $\mathcal{L}=||\mathbf{y}_{C1}-\mathbf{\hat x}_{C1}||_2$.
The conditioning is done by encoding the camera index $C_1$ into a trainable 3-dimensional vector that is then broadcast to match the $3\times128\times128$ size and concatenated with $\mathbf{t}$ to form the input of the network. Network $G_1$ uses ResNet50 U-net \cite{ronneberger2015unet,he2015deepresiduallearningimage} architecture with an additional sigmoid activation at the end. 
As it is apparent from the results in section \ref{sec:results}, the simplified synthetic reference $\mathbf{\hat{x}}_{C_1, L_2}$ yields only minor difference to the authentication performance of $\mathbf{\hat{x}}_{C_1,L_{comb}}$, thus we will make no distinction between them and denote them collectively as $\mathbf{\hat{x}}_{C_1}$ further in the text.

\subsection{Cross-camera dual-synthetic CDP translator}
The cross-camera dual-synthetic CDP translator is a network that conditionally transforms samples produced by imaging device $C_2$ into samples of imaging device $C_1$ while using the digital template information $\mathbf{t}$, namely, $\mathbf{\hat{x}}_{C2 \rightarrow C1}=G_2(\mathbf{t}, \mathbf{x}_{C2}, C_2, C_1)$. 
We parametrise this function by partitioning it into an encoder network $\mathbf{z}=E(\mathbf{x}_{C2}, C_2)$ that extracts the printer modulation from enrolled image $\mathbf{x}_{C2}$ and a decoder network $\mathbf{\hat{x}}_{C2 \rightarrow C1}=D(\mathbf{t}, \mathbf{z}, C_1)$ that applies this modulation to the template $\mathbf{t}$ to create an image that would mimic camera $C_1$. 
The encoder and decoder are optimised together using the $L_2$ loss $\mathcal{L}=||\mathbf{y}_{C1}-\mathbf{\hat{x}}_{C2 \rightarrow C1}||_2$. Both $E$ and $D$ have a ResNet50 U-net \cite{ronneberger2015unet,he2015deepresiduallearningimage} architecture with a sigmoid activation at the end, and are conditioned the same way as the synthetic generator $G_1$ in subsection \ref{subsec:synt}.

\subsection{Similarity and authentication performance}

The authentication is performed based on the metric of similarity between the reference $\bf r$ and the probe ${\bf y}_{C_1}$.
For the evaluation of the impact of reference $\bf r$ selection on the authentication accuracy, we subdivide each test CDP into non-overlapping $64\times64$ patches. We use networks $G_1$ and $G_2$ to generate references for each patch, with one camera acting as $C_1$ and another as $C_2$\footnote{To be noted that both cameras have different resolutions that impacts the accuracy of translation.}. In the result, we get 5 distinct references for camera $C_1$, namely:

\begin{itemize}\setlength\itemsep{0.1em}
    \item \em{Digital Template}: $\mathbf{t}$
    \item {Synthetic Reference}: $\mathbf{\hat{x}}_{C1}$
    \item {Dual-Synthetic Reference}: $\mathbf{\hat{x}}_{C2 \rightarrow C1}$
    \item {Physical Enrolment with a different camera}: \\
    $\mathbf{x}_{C2}, C_1\neq C_2$
    \item {Physical Enrolment with the same camera}: $\mathbf{x}_{C1}$
\end{itemize}

\begin{figure}[!htb]
    \centering
    \includegraphics[width=0.95\linewidth]{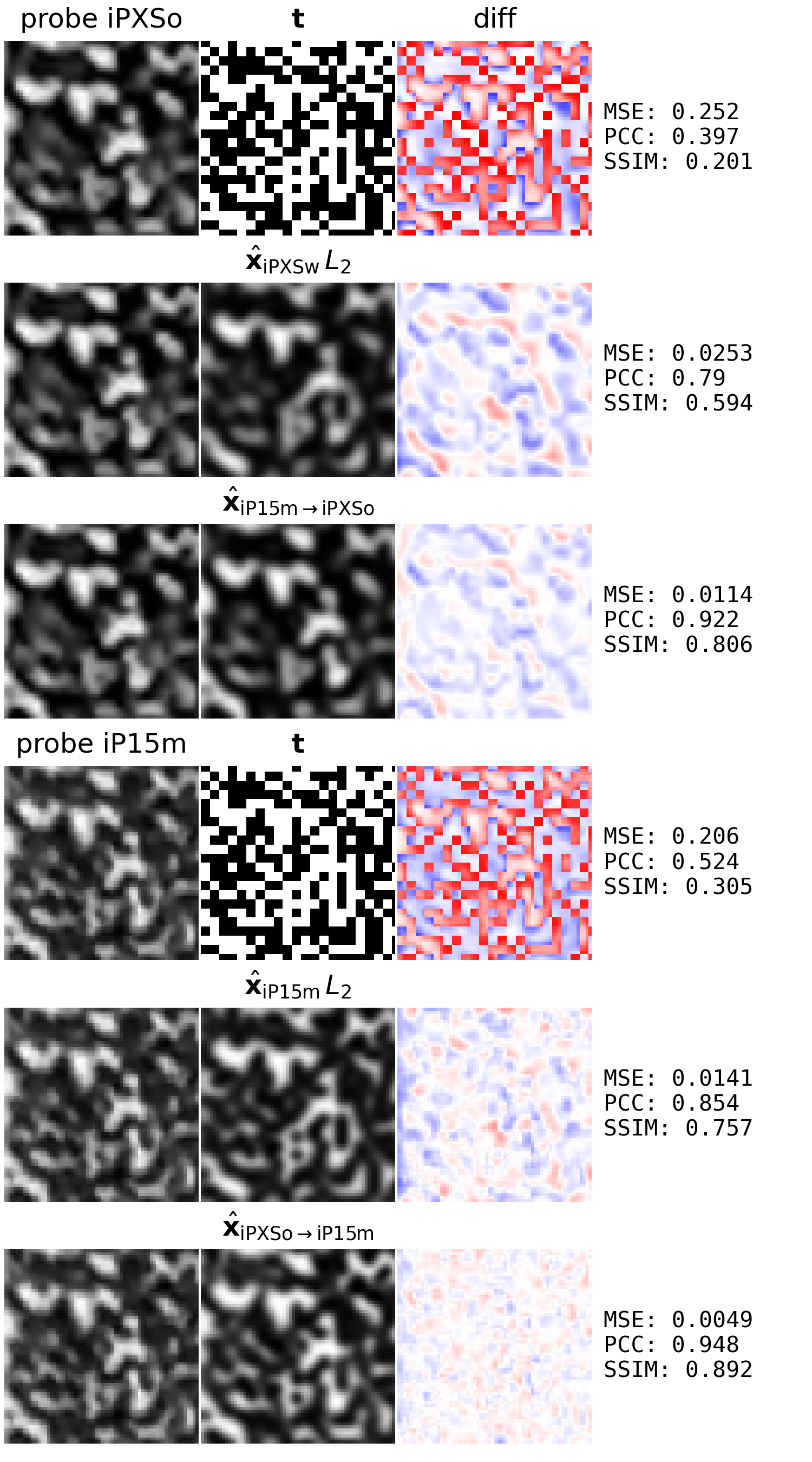}
    \caption{A small patch of a probe CDP from iPhone XS and iPhone 15 Pro Max images (first column), compared to digital template, synthetic reference and dual-synthetic reference (second column), with per-pixel difference (third column) and the similarity scores (fourth column).}
    \label{fig:example}
\end{figure}

An example of an original probe patch, its digital template, synthetic reference and dual-synthetic reference generated for this patch are shown in Fig. \ref{fig:example}.
We evaluate the similarity of each reference $\mathbf{r}$ with the probe $\mathbf{y}_{C1}\in\{\mathbf{y}_{C1}^{original}, \mathbf{y}_{C1}^{fake}\}$ using either mean square error (MSE), Pearson correlation coefficient (PCC), or structural similarity (SSIM) \cite{wang2004ssim} metrics.  
We use the evaluated scores to evaluate the Receiver Operating Characteristic (ROC) curve and find the Area Under the Curve (AUC) to define our authentication performance. 

\input{tables/combined}

%% file: tables/combined.tex
\begin{table*}[htbp]
\centering

\renewcommand{\arraystretch}{1.3}

\begin{tabular}{l|ccc|ccc|ccc|ccc}
\toprule
 & \multicolumn{6}{c|}{iPhone XS wide} & \multicolumn{6}{c}{iPhone 15 Pro macro} \\
 & \multicolumn{3}{c|}{Similarity} & \multicolumn{3}{c|}{AUC$\uparrow$} & \multicolumn{3}{c|}{Similarity} & \multicolumn{3}{c}{AUC$\uparrow$} \\
Reference & MSE$\downarrow$ & PCC$\uparrow$ & SSIM$\uparrow$ & MSE & PCC & SSIM & MSE$\downarrow$ & PCC$\uparrow$ & SSIM$\uparrow$ & MSE & PCC & SSIM \\
\midrule
$\mathbf{t}$ & 0.255 & 0.486 & 0.230 & 0.500 & 0.516 & 0.600 & 0.217 & 0.552 & 0.305 & 0.595 & 0.690 & 0.610 \\
$\mathbf{\hat{x}}_{C1,L_{comb}}$ \cite{chaban2024wifs} & 0.015 & 0.874 & 0.746 & 0.795 & 0.743 & 0.744 & 0.032 & 0.770 & 0.483 & 0.633 & 0.737 & 0.687 \\
$\mathbf{\hat{x}}_{C1,L_2}$ \cite{chaban2024wifs} & 0.013 & 0.889 & 0.753 & 0.828 & 0.772 & 0.786 & 0.015 & 0.872 & 0.737 & 0.758 & 0.816 & 0.770 \\
$\mathbf{\hat{x}}_{C2 \rightarrow C1}$ & \textbf{0.006} & \textbf{0.952} & \textbf{0.865} & \textbf{0.990} & \textbf{0.989} & \textbf{0.993} & \textbf{0.008} & \textbf{0.933} & \textbf{0.835} & \textbf{0.989} & \textbf{0.992} & \textbf{0.992} \\
$\mathbf{x}_{C2},C_2\neq C_1$ & 0.015 & 0.898 & 0.729 & 0.944 & 0.933 & 0.922 & 0.015 & 0.898 & 0.732 & 0.953 & 0.953 & 0.955 \\
\midrule
$\mathbf{x}_{C1}$ & 0.007 & 0.945 & 0.844 & 0.996 & 0.995 & 0.996 & 0.004 & 0.968 & 0.928 & 0.999 & 0.999 & 0.999 \\
\bottomrule
\end{tabular}
\caption{Average similarity metrics and ROC AUCs (original vs.\ fake discrimination) computed on blocks of $64\times64$ pixels ($21.3\times21.3$ digital template symbols) between the original images and various corresponding references. Within each device, similarity metrics are listed first, followed by the ROC AUCs based on those metrics. The arrow reflects superior performance for the similarity metrics. For each reference, $C_1$ denotes the device at the top of the column and $C_2$ denotes the device from the opposite side of the table.}
\label{tab:sim_auc}
\end{table*}

%% file: sections/Section-IT.tex
\section{Information-Theoretic Justification}
\label{sec:info}

We justify why using \emph{both} the digital template $\mathbf{t}$ and an enrolled physical capture $\mathbf{x}_{C2}$ yields a strictly more informative reference for the verification capture $\mathbf{x}_{C1}$ than using $\mathbf{t}$ alone, and why a learned dual-synthetic mapping based on $(\mathbf{t},\mathbf{x}_{C2})$ improves authentication.

Firstly, let all random variables be defined on a common probability space. The chain rule gives us a decomposition of mutual information \cite{CoverInfo}:
\begin{align}
    I(\mathbf{X}_{C1}; \mathbf{T}, \mathbf{X}_{C2}) &= I(\mathbf{X}_{C1}; \mathbf{T}) + I(\mathbf{X}_{C1}; \mathbf{X}_{C2} | \mathbf{T}) \\
    & \geq I(\mathbf{X}_{C1}; \mathbf{T}),
\end{align}
due to the fact that $I(\mathbf{X}_{C1}; \mathbf{X}_{C2} \mid \mathbf{T}) \geq 0$. 
Thus, \emph{on average}, the dual-reference $(\mathbf{T},\mathbf{X}_{C2})$ is strictly more informative for the verification than the template $\mathbf{T}$ alone.

Secondly, let $\widehat{\mathbf{X}}_{C_1} = \mathbb{E}[\mathbf{X}_{C1}| \mathbf{T}]$ and 
$\widehat{\mathbf{X}}_{C_2\rightarrow C_1} \; = \mathbb{E}[\mathbf{X}_{C1}| \mathbf{T},\mathbf{X}_{C2}]$ 
denote the $L_2$-optimal predictors of $\mathbf{X}_{C1}$ using the single- and dual-reference information, respectively, and $ \mathbb{E}[.]$ denotes a mathematical expectation. The \emph{law of total variance}  \cite{blitzstein2014introduction} implies the matrix PSD ordering
\begin{equation}
  \mathrm{Cov}(\mathbf{X}_{C1}\mid \mathbf{T},\mathbf{X}_{C2})
  \;\preceq\;
  \mathrm{Cov}(\mathbf{X}_{C1}\mid \mathbf{T}).
  \label{eq:cov_order}
\end{equation}
The minimum mean squared error (MMSE) defined as $\operatorname{mmse}(\mathbf{X} \mid \mathbf{Y})=\mathbb{E}\left[\|\mathbf{X}-\mathbb{E}[\mathbf{X} \mid \mathbf{Y}]\|^2_2\right]$ leads to inequality  $\mathrm{mmse}(\mathbf{X}_{C1}\mid \mathbf{T},\mathbf{X}_{C2}) \le \mathrm{mmse}(\mathbf{X}_{C1}\mid \mathbf{T})$.
Consequently, for networks trained with an $L_2$ loss, one has:
\begin{equation}
  \mathbb{E}\bigl[\|\mathbf{X}_{C1} - \widehat{\mathbf{X}}_{C_2\rightarrow C_1} \|_2^2\bigr]
  \;\le\;
  \mathbb{E}\bigl[\|\mathbf{X}_{C1} - \widehat{\mathbf{X}}_{C_1}\|_2^2\bigr].
  \label{eq:mmse_order}
\end{equation}
Since common similarity scores (negative MSE, PCC, SSIM in the small-error regime) are monotone in the prediction error, \eqref{eq:mmse_order} predicts higher similarity between the dual-synthetic reference and the probe, matching the empirical results.

%% file: sections/Secrion-results.tex
\section{Results}
\label{sec:results}

\subsection{Image similarity}

\begin{figure*}[t]
    \centering
    \begin{subfigure}{.49\textwidth}
        \includegraphics[width=\textwidth]{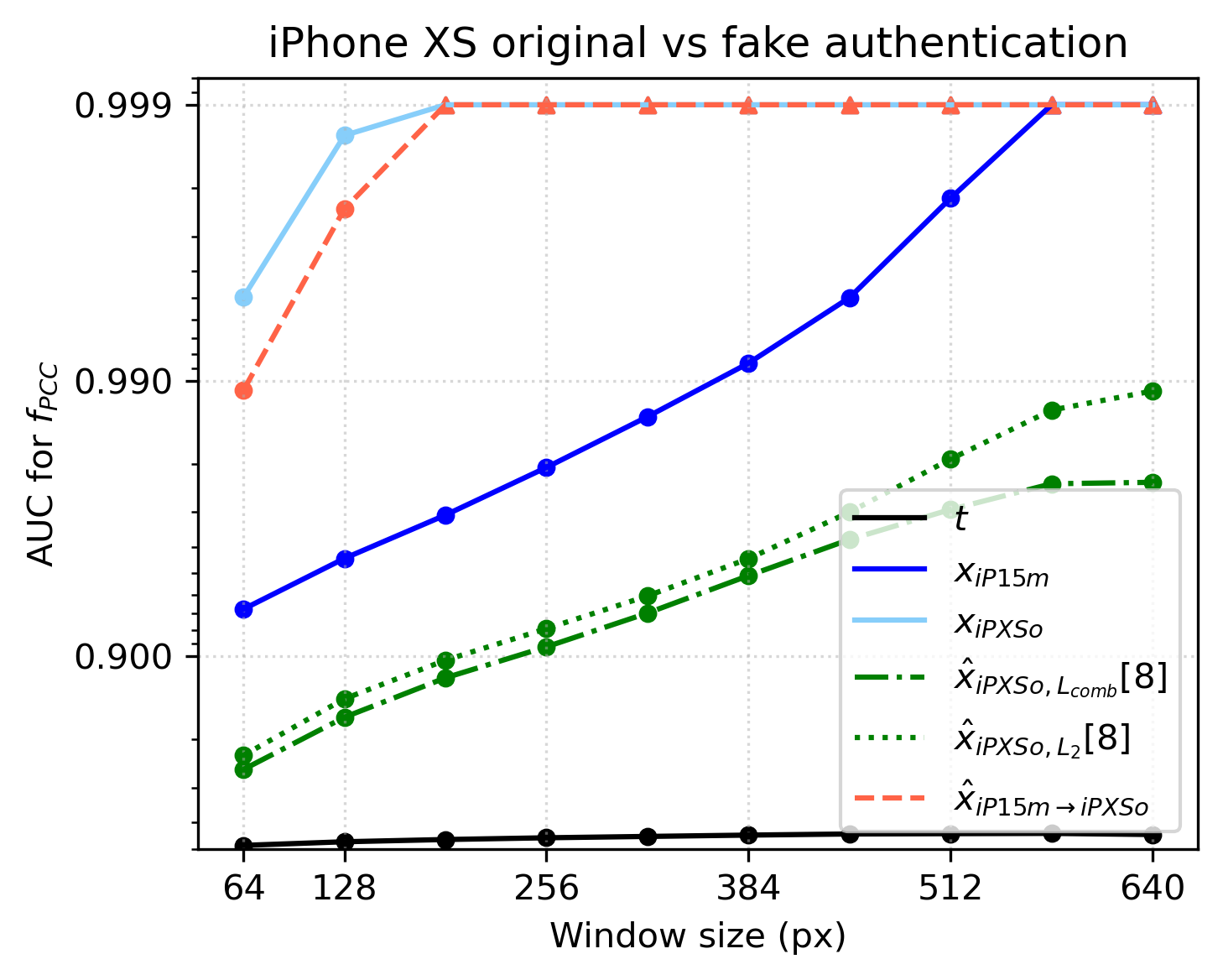}
        \caption{Low-end verification camera}
        \label{fig:auc_vs_size_iPXS}
    \end{subfigure}
    \begin{subfigure}{.49\textwidth}
        \includegraphics[width=\textwidth]{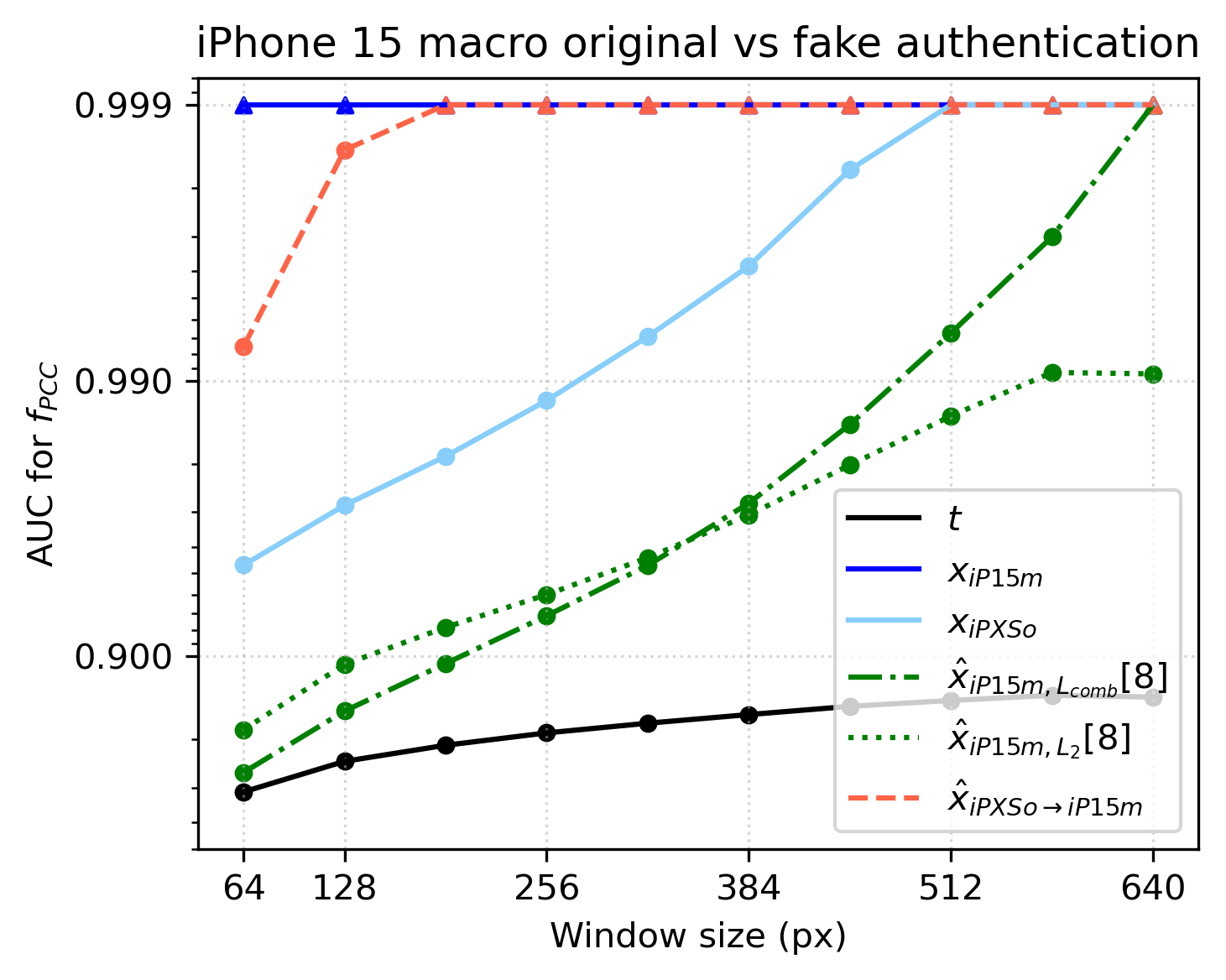}
        \caption{High-end verification camera}
        \label{fig:auc_vs_size_iP15s}
    \end{subfigure}
    \caption{ROC AUC of authentication performance depending on the window size, shown for 5 different references used. The values are clipped at 0.999 as the given statistic does not allow for higher precision estimation.}
    \label{fig:roc_main}
\end{figure*}

Table~\ref{tab:sim_auc} reports average similarity metrics between references and original images.
The bottom row shows the idealised case where reference and evaluation cameras coincide; since images were taken consecutively under identical conditions, differences are limited to camera noise, minor parameter auto-adjustment (e.g., dynamic focusing, ISO, etc.) and uncertainty in geometrical image alignment.
Here, consecutive iPhone 15 macro images differ less than iPhone XS wide-angle images, reflecting the XS’s lower resolution.
The top row confirms this: iPhone 15 Pro macro images yield much higher similarity with the digital template than XS.
When using a different camera as reference ($\mathbf{x}_{C2}, C_1\neq C_2$), similarity drops, though printer variations remain captured.
Synthetic reference ($\mathbf{\hat{x}}_{C1}$) reproduces the correct camera “style” but misses stochastic printing effects, while dual-synthetic reference ($\mathbf{\hat{x}}_{C2\rightarrow C1}$) combines both, achieving the best similarity scores among all realistic methods. This effect is also visible in Fig. \ref{fig:example}, where $\mathbf{\hat{x}}_{C1}$ shows several regions of erroneous reconstruction, while $\mathbf{\hat{x}}_{C2\rightarrow C1}$ corrects these details with the information from the enrolment.

To construct the $\mathbf{\hat{x}}_{15 \rightarrow XS}$ reference, we use iPhone 15 Pro macro enrolment captures, which have higher resolution details than those from the XS. 
This explains why the similarity of $\mathbf{\hat{x}}_{15 \rightarrow XS}$ to $\mathbf{y}_{XS}$ can exceed that between two noisy XS images ($\mathbf{x}_{XS}$, $\mathbf{y}_{XS}$).
In the $\mathbf{\hat{x}}_{XS \rightarrow 15}$ case, we use a lower-quality enrolment camera to produce a reference for a higher-quality verification camera. 
Although the gain is not as high as in the $\mathbf{\hat{x}}_{15 \rightarrow XS}$ case due to limited information, it still shows that even poor enrolment quality can yield significant similarity improvements due to the captured printed CDP stochasticity.

\subsection{Authentication on small blocks}

Authentication performance measured by the area under the curve (AUC) of the receiver operating characteristic for different references is presented in Table~\ref{tab:sim_auc}.
As expected, better similarity metrics generally correspond to higher AUCs: the closer a reference is to the original, the more effectively differences between genuine and fake images can be detected. 

Consistent with \cite{chaban2024wifs}, consecutive captures achieve near-perfect performance across all devices, while digital templates perform poorly, particularly for lower-end models. 
Adapting the template to match a device’s imaging style ($\mathbf{\hat{x}}_{C1}$) improves results, but remains limited since it only reflects mismodeling of the template and ignores printer-specific variations. By contrast, cross-camera referencing ($\mathbf{x}_{C2}, C_1\neq C_2$) performs better despite similar similarity scores, as it captures both template decoding errors and printer-induced discrepancies. Finally, the dual-synthetic reference ($\mathbf{\hat{x}}_{C2\rightarrow C1}$) combines these advantages—replicating device style while incorporating printing artefacts—yielding the best authentication performance among realistic methods.

\subsection{Block Aggregation}

One way to enhance the security of a CDP is to increase the number of symbols in the digital template, i.e., enlarge its size, since this increases the total number of decoding errors an attacker is likely to make in mean when decreasing the variance according to the concentration property for Hamming weight.
To evaluate similarity within a probe window, we compute scores on $64\times64$ blocks and then average the block scores $s_i$ within a larger $64N\times64N$ window to obtain the window score $S = \frac{1}{N} \sum_{i=1}^N s_i$.

Figure~\ref{fig:roc_main} shows that authentication performance based on PCC improves with larger window sizes (with MSE and SSIM yielding analogous results). Importantly, the relative ranking of reference qualities remains consistent across window sizes, with the dual-synthetic reference maintaining a substantial advantage over other realistic cases.
This indicates that the dual-synthetic reference enables more robust large-CDP authentication than prior methods, including \cite{chaban2024wifs}. At the same time, our method achieves AUC $\approx$ 0.999 using only 9\% of the full CDP area.
Such reliable authentication from a fraction of the CDP is particularly valuable when parts of it are damaged, obscured by glare, or when a more compact design is desired.

In the Supplementary Material, we provide plots analogous to Figure~\ref{fig:roc_main} for 7 imaging devices available in the dataset (Epson scanner, iPhone XS, iPhone 12, iPhone 14 Pro Max main and macro cameras, iPhone 15 Pro Max main and macro cameras). The results confirm that the trends summarised in this section generalise to a wide range of imaging devices.

%% file: sections/Section-conclusion.tex
\section{Conclusion}
\label{sec:conclusion}

We introduced a cross-camera dual-synthetic referencing strategy \footnote{The codebase will be available on GitHub upon publication \url{https://github.com/IvanOleksiyuk/cdp-dual-synthetics/}.} that substantially improves CDP authentication accuracy, especially under challenging imaging conditions and when using small patch sizes. Grounded in mutual information theory and validated across multiple imaging devices, our method leverages print-specific stochastic variations that remain unattainable for attackers, even with state-of-the-art template decoding. Although reliable CDP authentication (AUC$>$0.99) in \cite{chaban2024wifs} was achieved only for high-end cameras such as the iPhone 15 Pro Max, by introducing the dual-synthetic reference, we enable effective CDP authentication for devices previously considered unsuitable, like iPhone XS.

By enabling reliable authentication from small CDP blocks, the approach is inherently robust to damage and glare, as the corrupted regions can simply be excluded, and may support faster verification by transmitting only a fraction of the CDP. Although per-pattern enrolment is required, each CDP needs to be enrolled with only a single device. Moreover, we show that even lower-quality devices contribute valuable information, lowering the cost of enrolment equipment. Additionally, the training of the model took only 51 minutes using one commercial-grade $\text{NVIDIA}^{\text{\tiny{®}}}$ RTX 2080 Ti GPU, and the inference for one CDP takes on average 0.66 s with the same hardware, making the method attractive for real-world applications.

In this work, we directly use the reference–probe similarity distance as the discriminative score, resulting in an unsupervised counterfeit detection approach that remains unbiased toward specific counterfeit implementations. In future work, this method could be further extended by combining these distances to construct either an unsupervised \cite{chaban2021fakes, app13148101} (one-class) or a supervised \cite{computers12090183,chaban2021fakes} (two-class) classifier.

Future work will explore domain adaptation and zero-shot generalisation, further extending the practicality and resilience of CDP-based authentication.

%% file: sections/Section-multi-camera.tex
\section{Extended imaging device case study}
\label{sec:intro}

\begin{figure*}[ht]
    \centering
    \includegraphics[width=\textwidth]{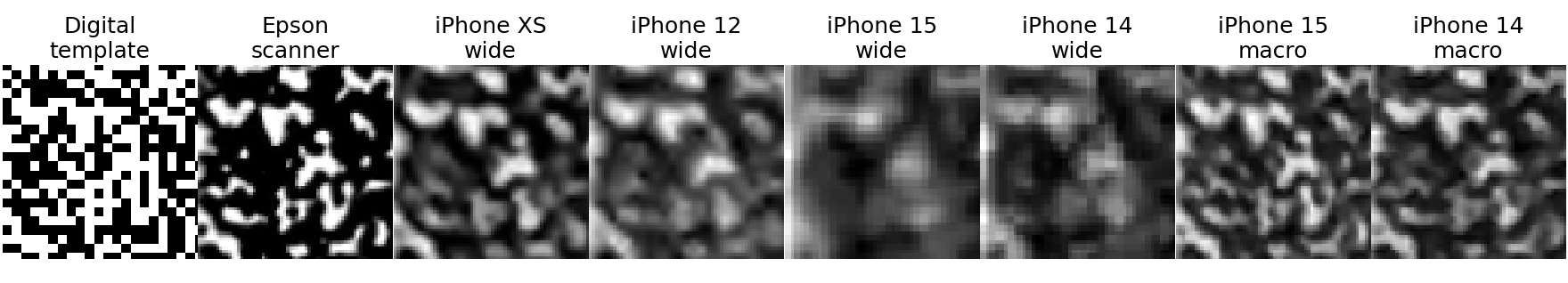}
    \caption{A patch of digital template of a CDP and the captures of the same patch by 7 devices available in the dataset.}
    \label{fig:camera-demo}
\end{figure*}

In the main text, to keep the presentation concise, we focused on two representative imaging devices spanning different performance levels: the iPhone XS main camera as a lower-end imaging device and the iPhone 15 Pro Max macro (ultra-wide-angle) camera as a higher-end imaging device. In this Supplementary Material, we provide additional results demonstrating that the conclusions drawn in the main text can be extended to a wide range of imaging devices.

To this end, we train both the synthetic reference generator $G_1$ and the dual-synthetic reference generator $G_2$ using conditioning on seven classes of imaging devices available in the dataset, namely:
\begin{itemize}
\item Epson V850 digital scanner
\item iPhone XS main camera
\item iPhone 12 main camera
\item iPhone 14 Pro Max main camera
\item iPhone 15 Pro Max main camera
\item iPhone 14 Pro Max macro camera
\item iPhone 15 Pro Max macro camera.
\end{itemize}

An example patch of one CDP is shown for each camera in Fig. \ref{fig:camera-demo}

\begin{figure*}[t]
    \centering
    \begin{subfigure}{.49\textwidth}
        \includegraphics[width=\textwidth]{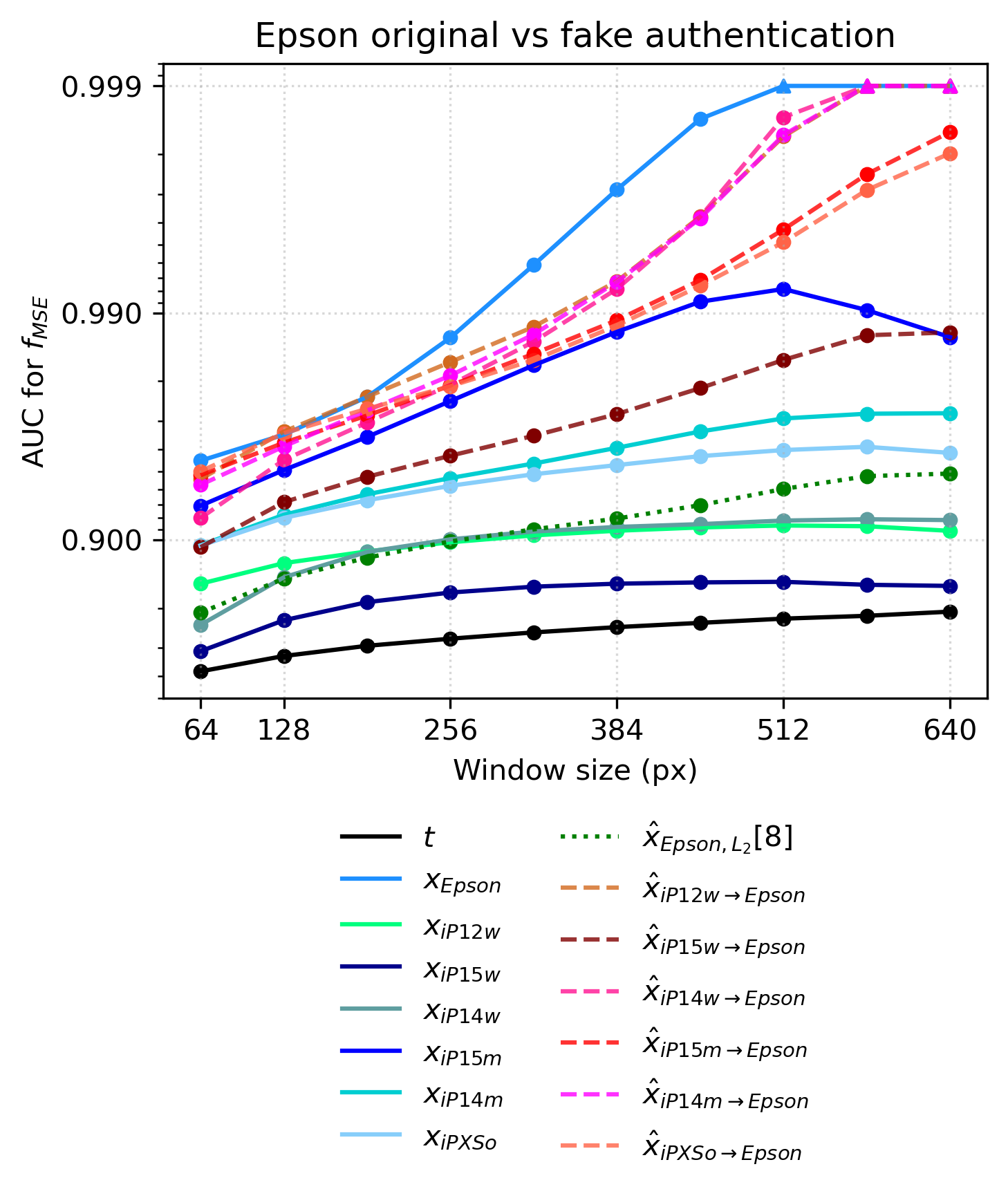}
        \caption{Separate curves}
        \label{fig:separate}
    \end{subfigure}
    \begin{subfigure}{.49\textwidth}
        \includegraphics[width=\textwidth]{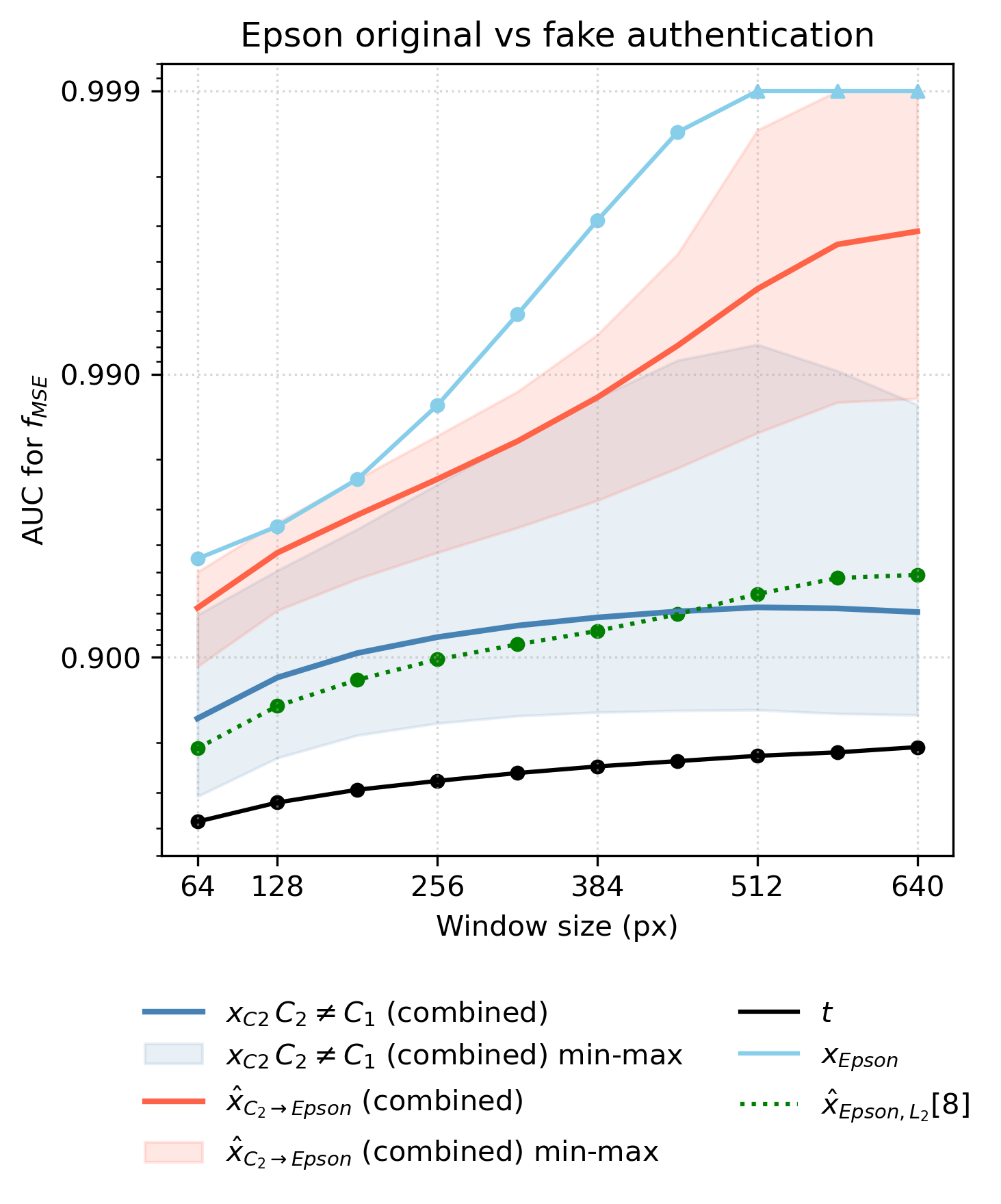}
        \caption{Combined curves}
        \label{fig:combined}
    \end{subfigure}
    \caption{ROC AUC of authentication performance depending on the window size, shown for different references used. The values are clipped at 0.999 as the given statistic does not allow for higher precision estimation.}
    \label{fig:roc1}
\end{figure*}

For these models, we report 21 plots (corresponding to seven verification cameras and three metrics), analogous to Fig. 3 in the main text. These plots are given in "separate curves" subplots of figures below e.g. Fig. \ref{fig:separate}. These plots include the performance of all reference types: the digital template $\mathbf{t}$, the synthetic reference $\hat{\mathbf{x}}_{C1,L_2}$, the dual-synthetic reference $\hat{\mathbf{x}}_{C2 \rightarrow C1}$ for each enrolment camera $C_2$, the real reference $\mathbf{x}_{C2}$ for all $C_2 \neq C_1$, and the same-camera reference $\mathbf{x}_{C1}$.

In addition, to improve visual clarity, we provide another set of 21 analogous plots, in which all dual-synthetic references and different-camera references are aggregated by reporting their average performance as a solid line, together with the corresponding min–max range shown as a shaded band. These plots are given in "combined curves" subplots of figures below e.g. Fig. \ref{fig:separate}.

The main observation and conclusion of the main text is that the dual-synthetic reference $\hat{\mathbf{x}}_{C2 \rightarrow C1}$ yields better authentication performance than both the synthetic reference $\hat{\mathbf{x}}_{C1,L_2}$ and the digital reference $\mathbf{t}$. Examination of all 21 plots confirms that this conclusion holds consistently across all enrolment and verification cameras, as well as for all window sizes.

In addition, in most cases, the dual-synthetic reference $\hat{\mathbf{x}}_{C2 \rightarrow C1}$ also outperforms a real reference image $\mathbf{x}_{C2}$ acquired from an enrolment camera $C_2 \neq C_1$ that differs from the verification camera $C_1$. This trend holds except when $C_2$ and $C_1$ correspond to very similar devices, for example, the corresponding cameras of the iPhone 14 Pro Max and iPhone 15 Pro Max.

We use the same hyperparameters (learning rate, maximum number of training epochs, and model size) for this seven-class setting as in the two-class setting. Because the model must accommodate a larger number of device classes, the quantitative performance for each class—including the original two classes—is marginally reduced compared to the two-class case. However, this does not affect the qualitative ranking of reference types or the overall conclusions regarding authentication performance. We expect that re-optimizing the hyperparameters for the seven-class model would allow it to recover performance comparable to that of the original two-class model.


\begin{figure*}[t]
    \centering
    \begin{subfigure}{.49\textwidth}
        \includegraphics[width=\textwidth]{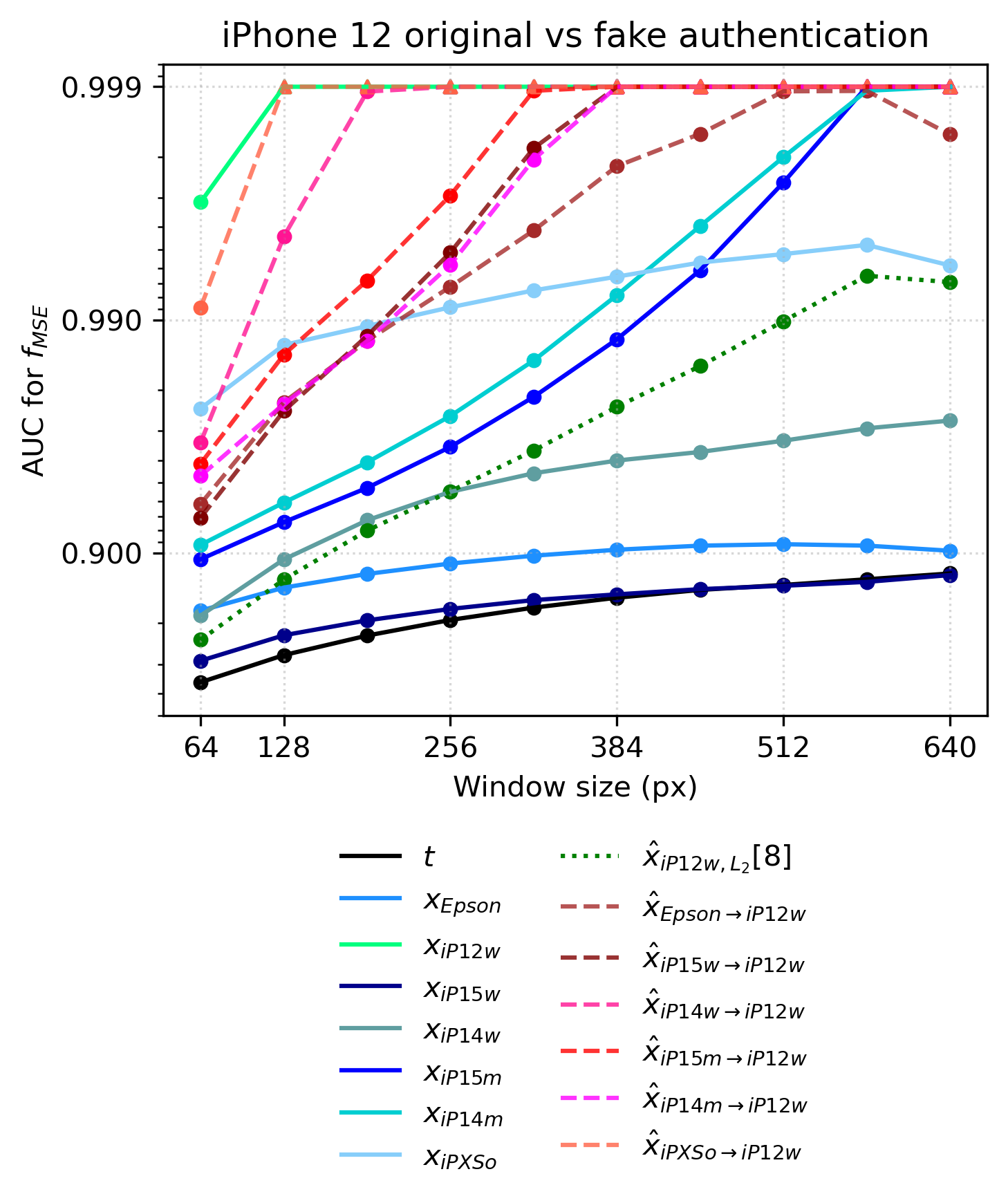}
        \caption{Separate curves}
    \end{subfigure}
    \begin{subfigure}{.49\textwidth}
        \includegraphics[width=\textwidth]{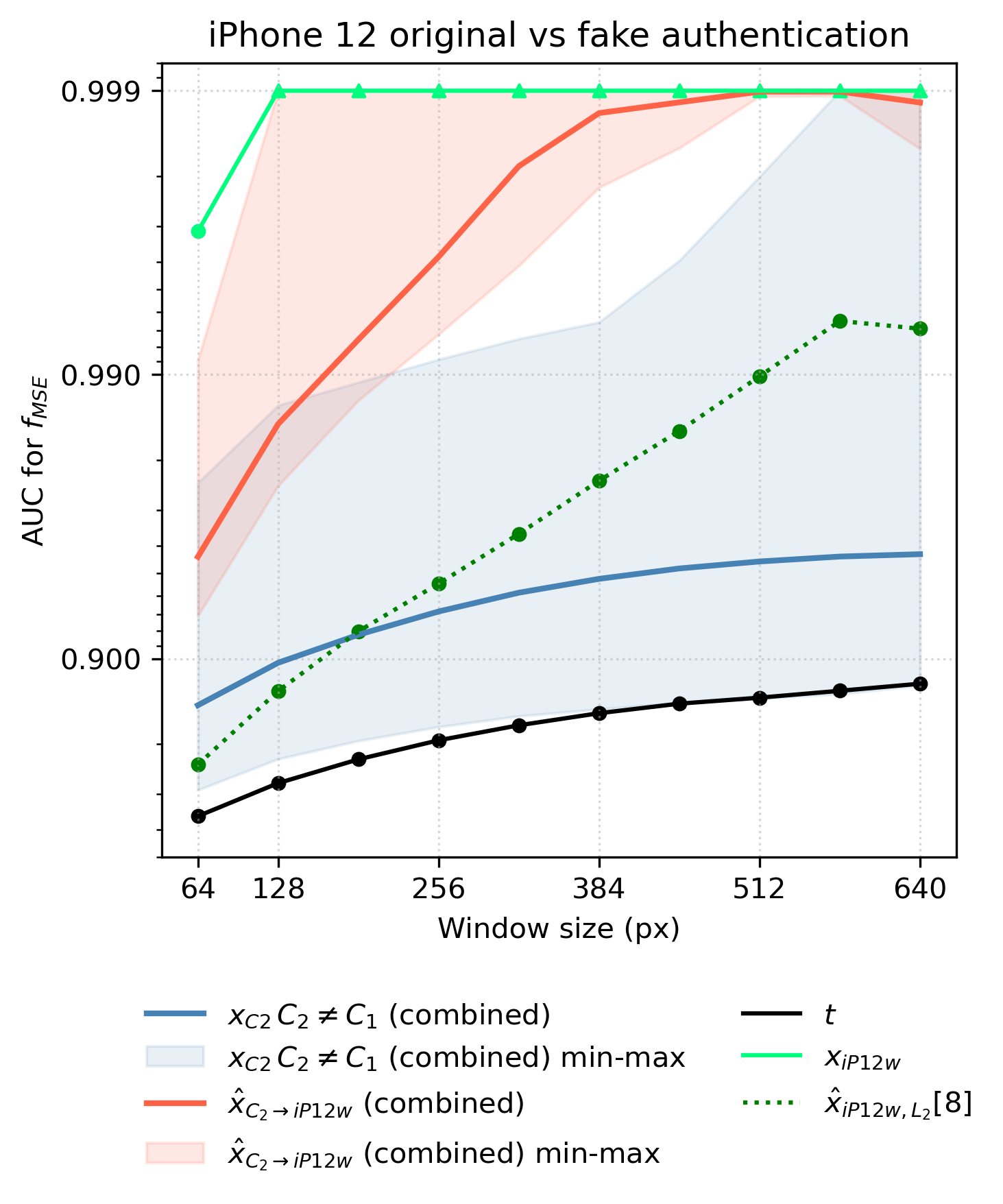}
        \caption{Combined curves}
    \end{subfigure}
\end{figure*}

\begin{figure*}[t]
    \centering
    \begin{subfigure}{.49\textwidth}
        \includegraphics[width=\textwidth]{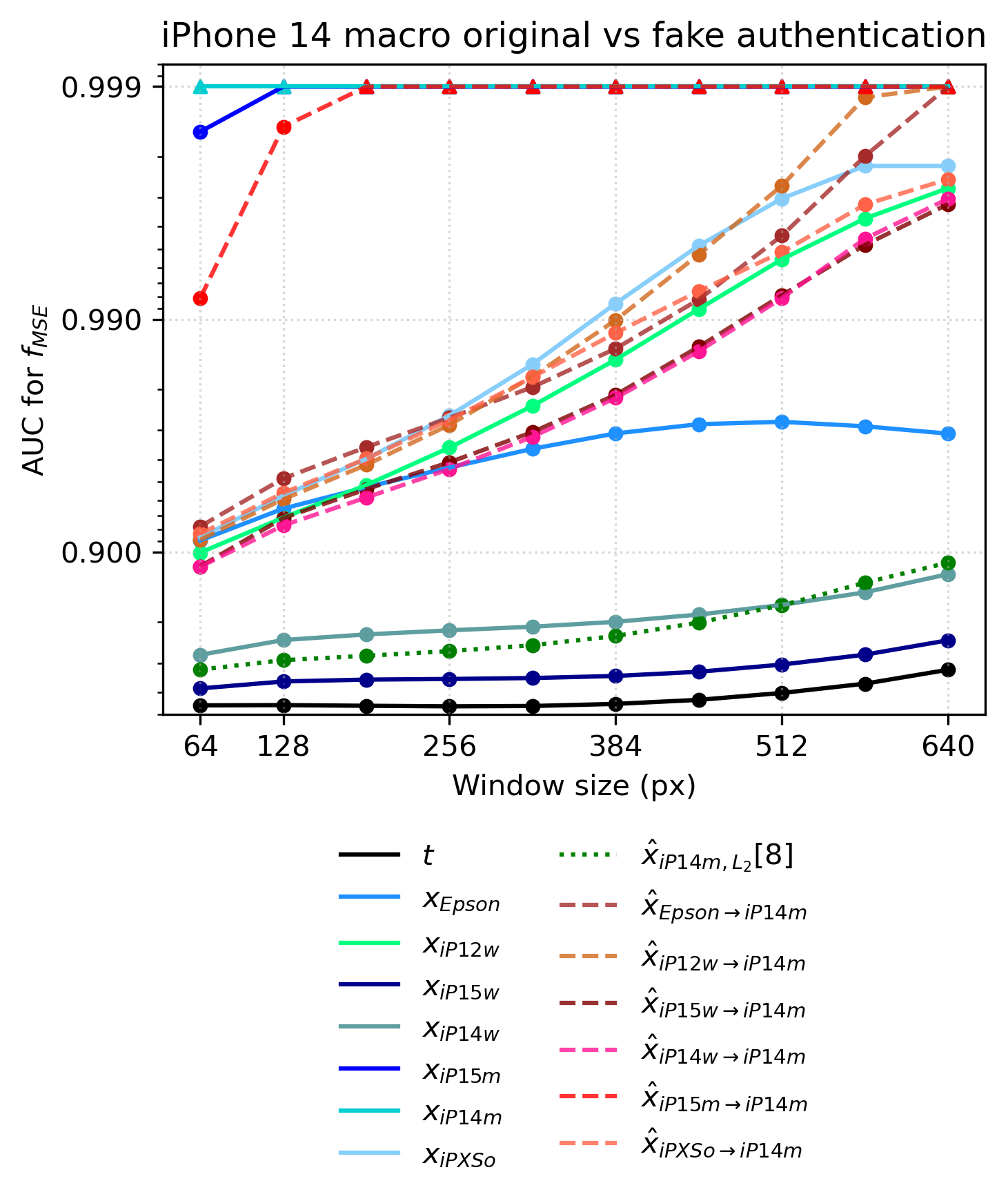}
        \caption{Separate curves}
    \end{subfigure}
    \begin{subfigure}{.49\textwidth}
        \includegraphics[width=\textwidth]{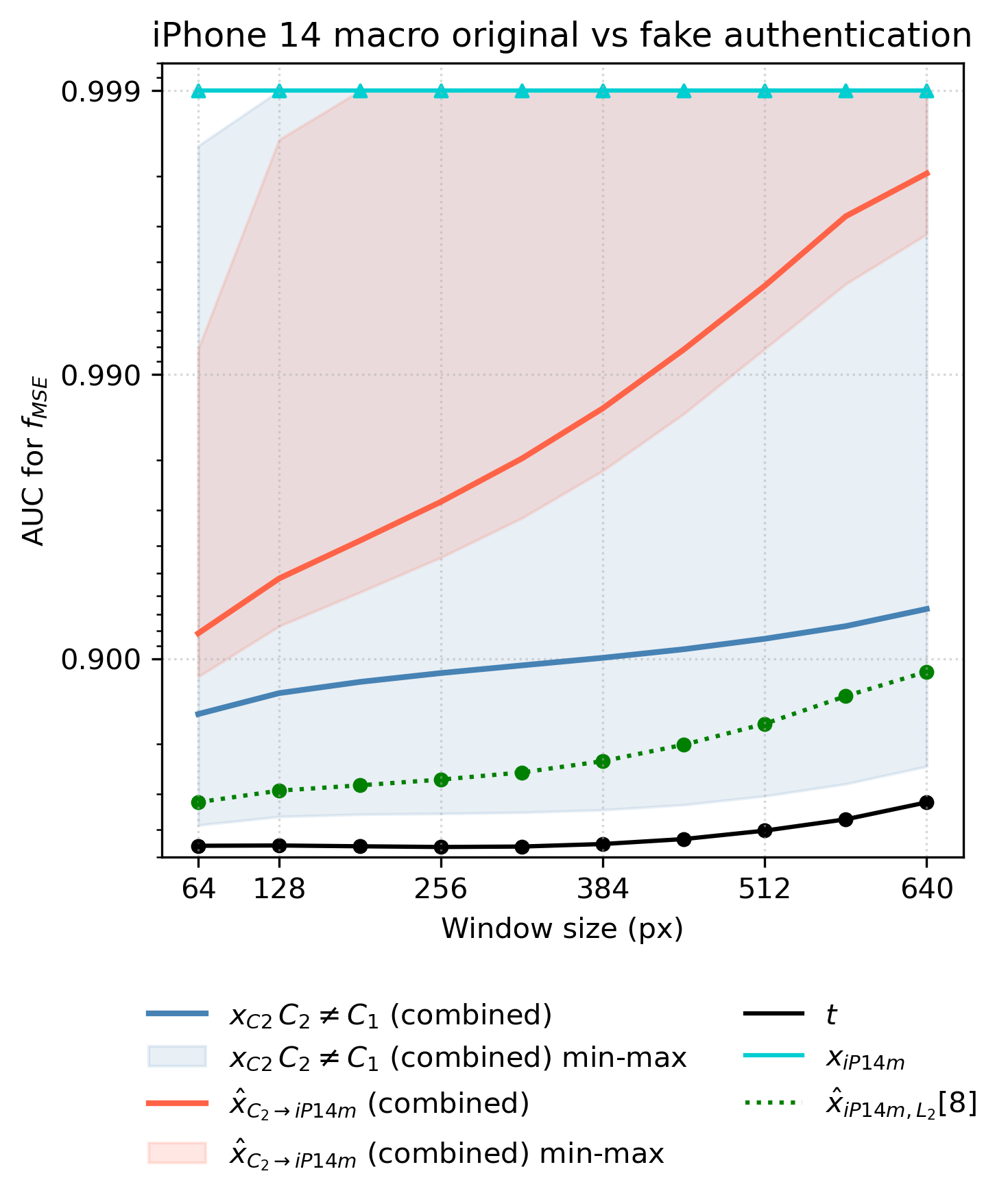}
        \caption{Combined curves}
    \end{subfigure}
\end{figure*}

\begin{figure*}[t]
    \centering
    \begin{subfigure}{.49\textwidth}
        \includegraphics[width=\textwidth]{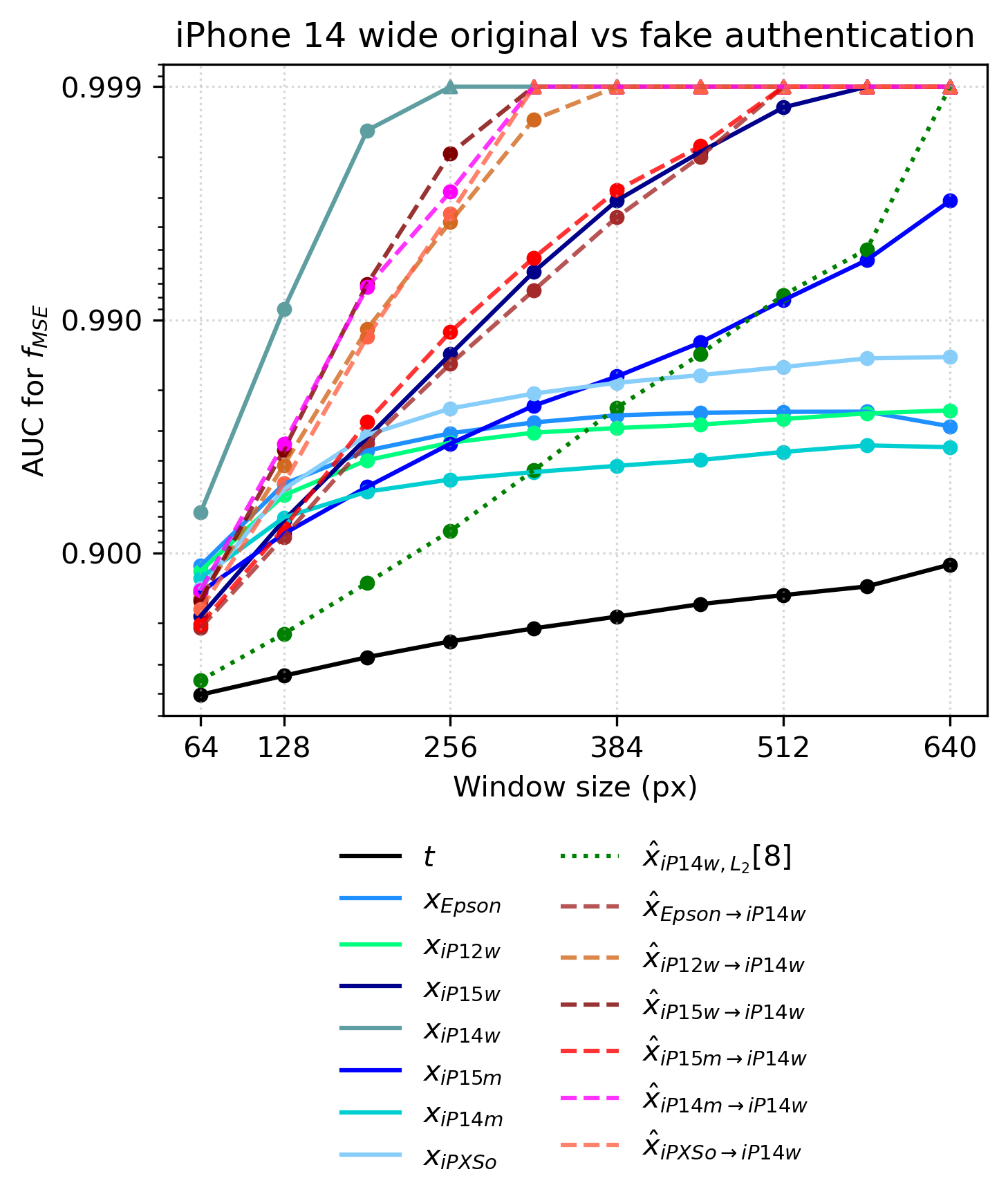}
        \caption{Separate curves}
    \end{subfigure}
    \begin{subfigure}{.49\textwidth}
        \includegraphics[width=\textwidth]{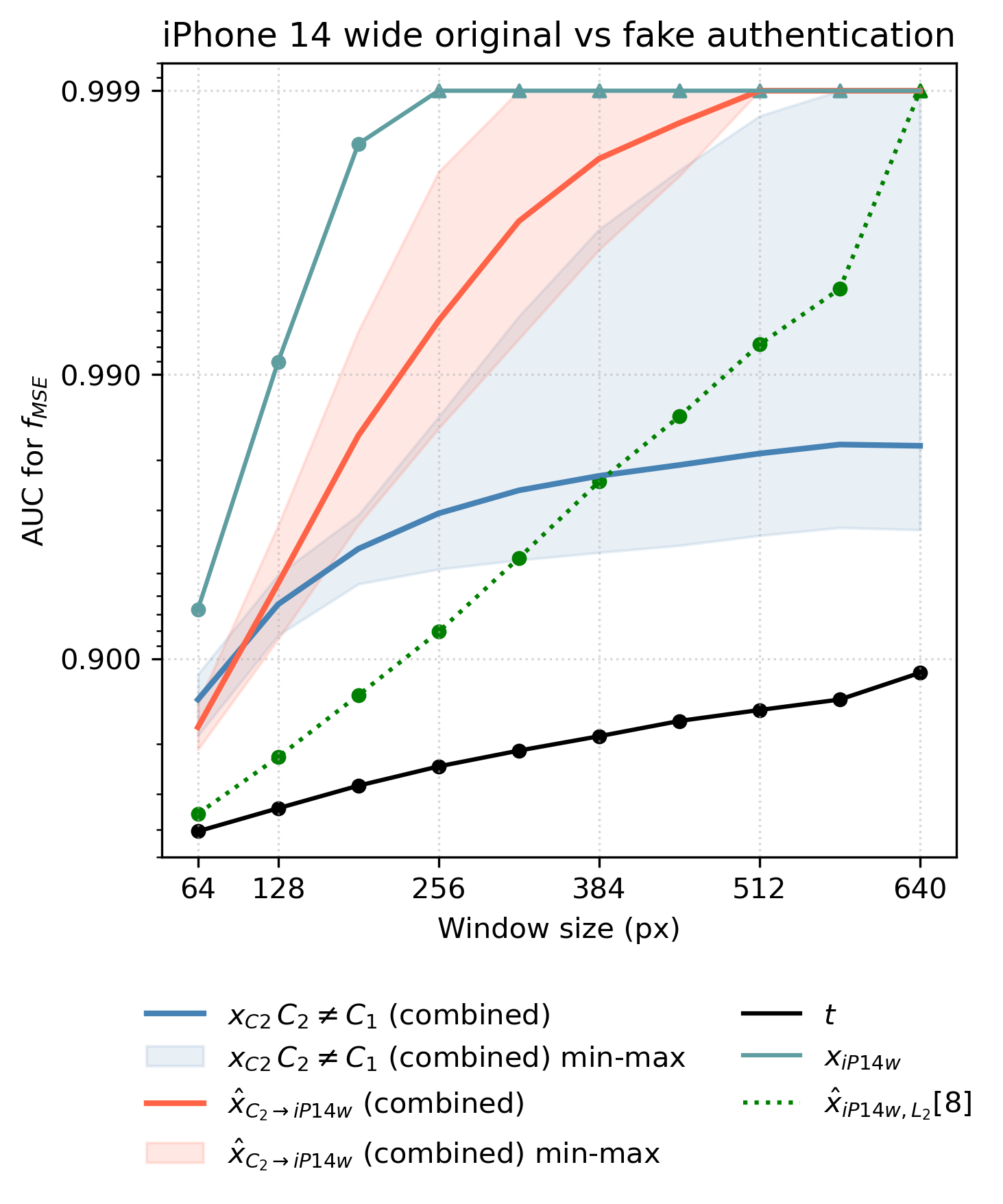}
        \caption{Combined curves}
    \end{subfigure}
\end{figure*}

\begin{figure*}[t]
    \centering
    \begin{subfigure}{.49\textwidth}
        \includegraphics[width=\textwidth]{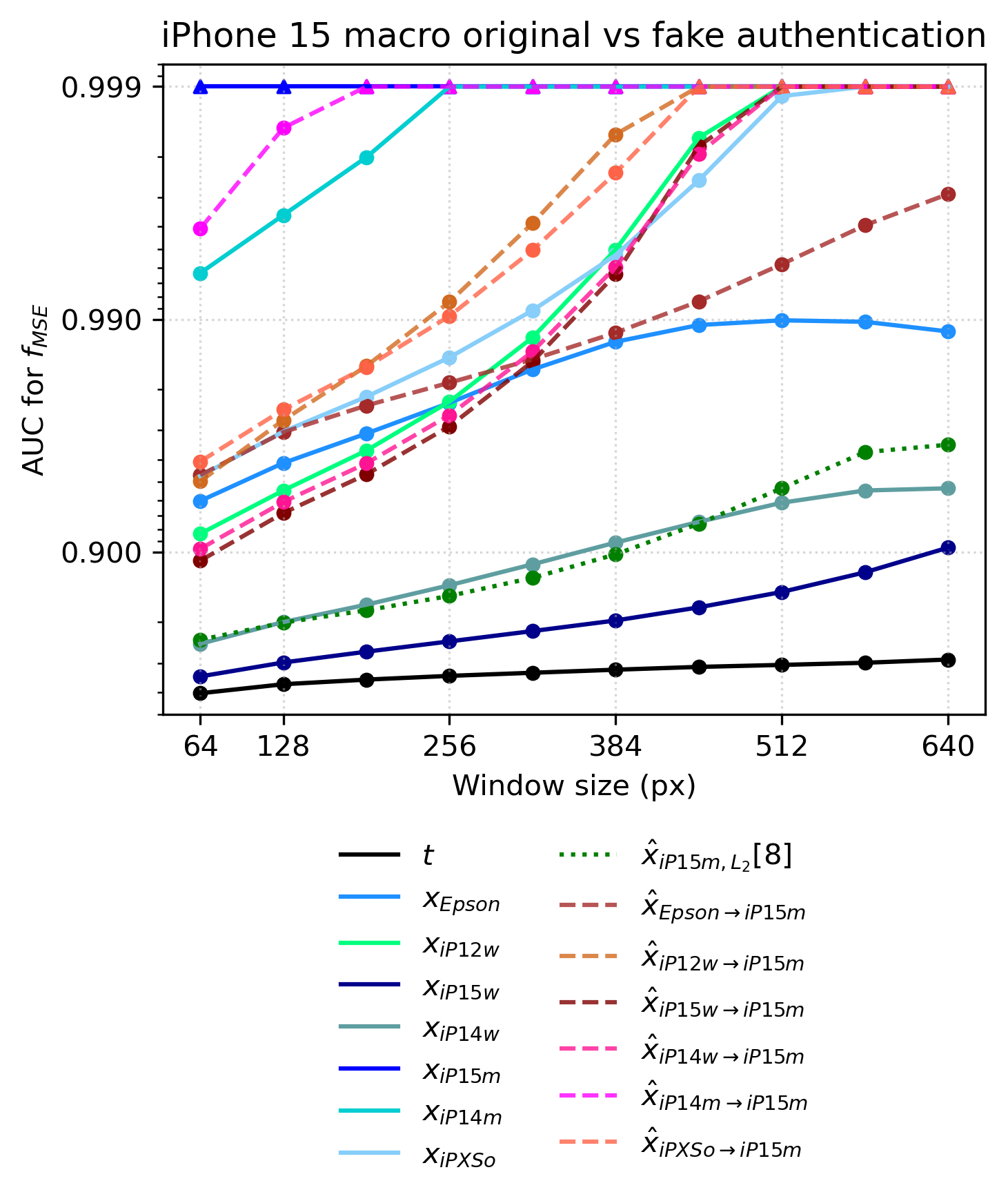}
        \caption{Separate curves}
    \end{subfigure}
    \begin{subfigure}{.49\textwidth}
        \includegraphics[width=\textwidth]{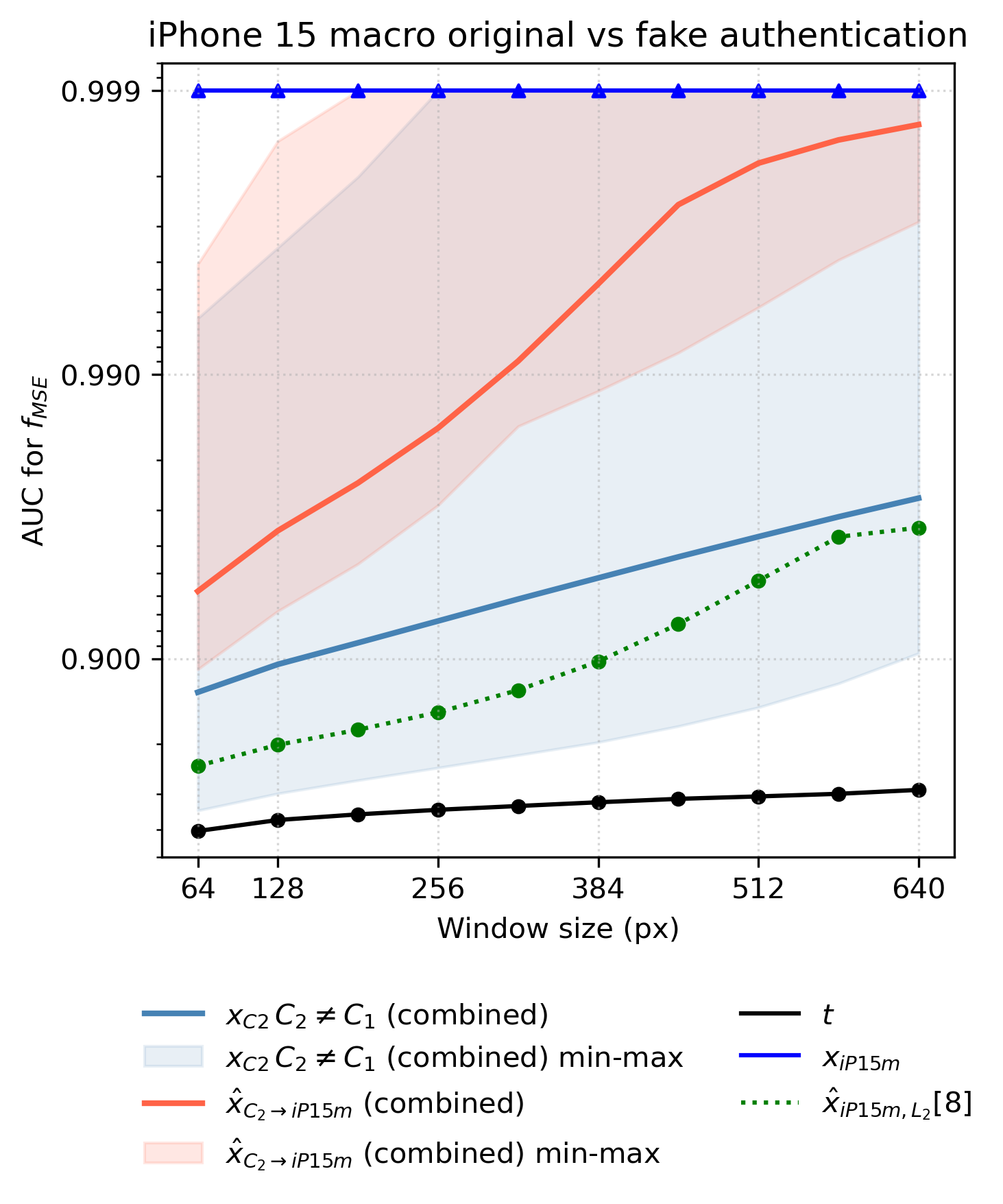}
        \caption{Combined curves}
    \end{subfigure}
\end{figure*}

\begin{figure*}[t]
    \centering
    \begin{subfigure}{.49\textwidth}
        \includegraphics[width=\textwidth]{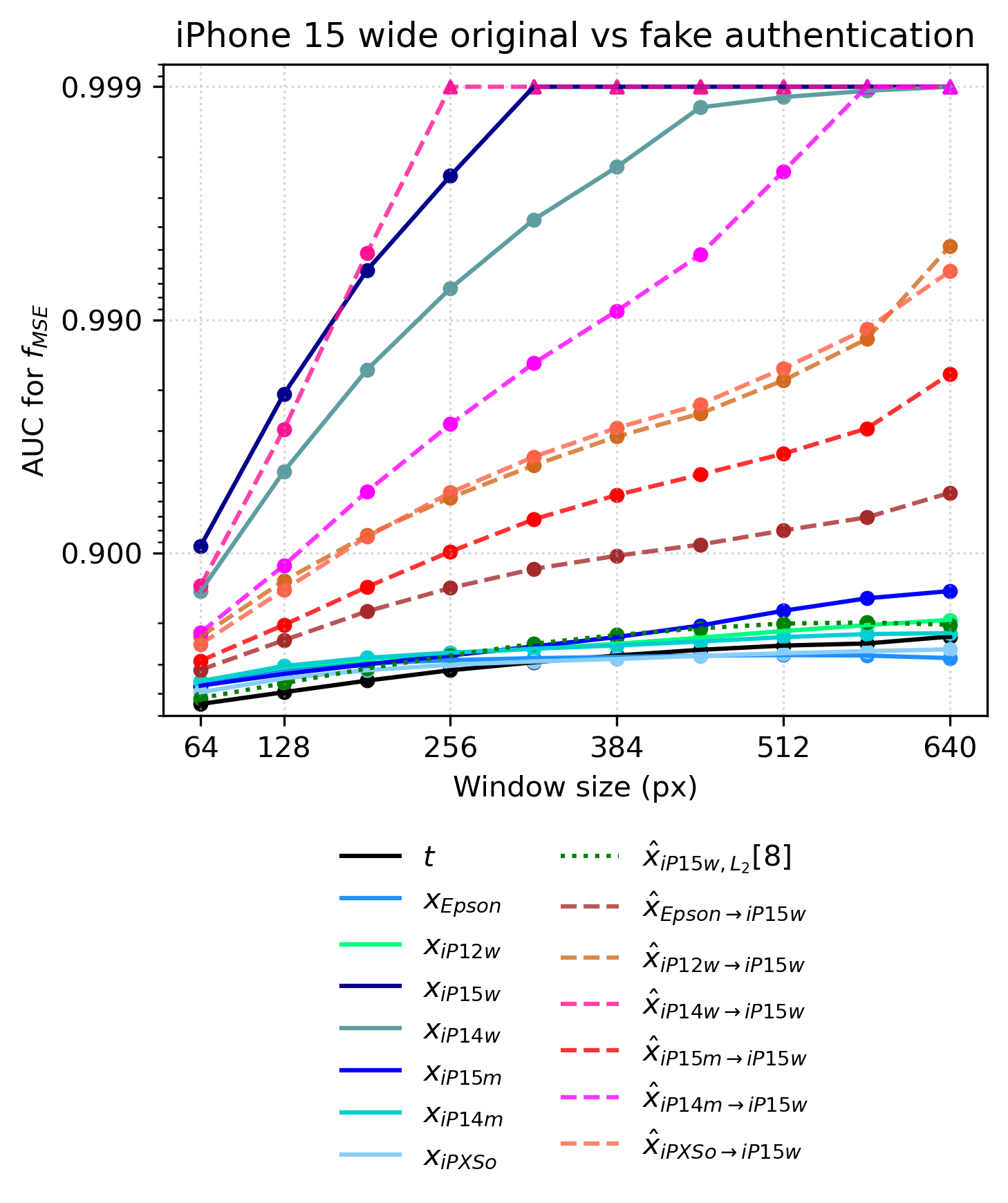}
        \caption{Separate curves}
    \end{subfigure}
    \begin{subfigure}{.49\textwidth}
        \includegraphics[width=\textwidth]{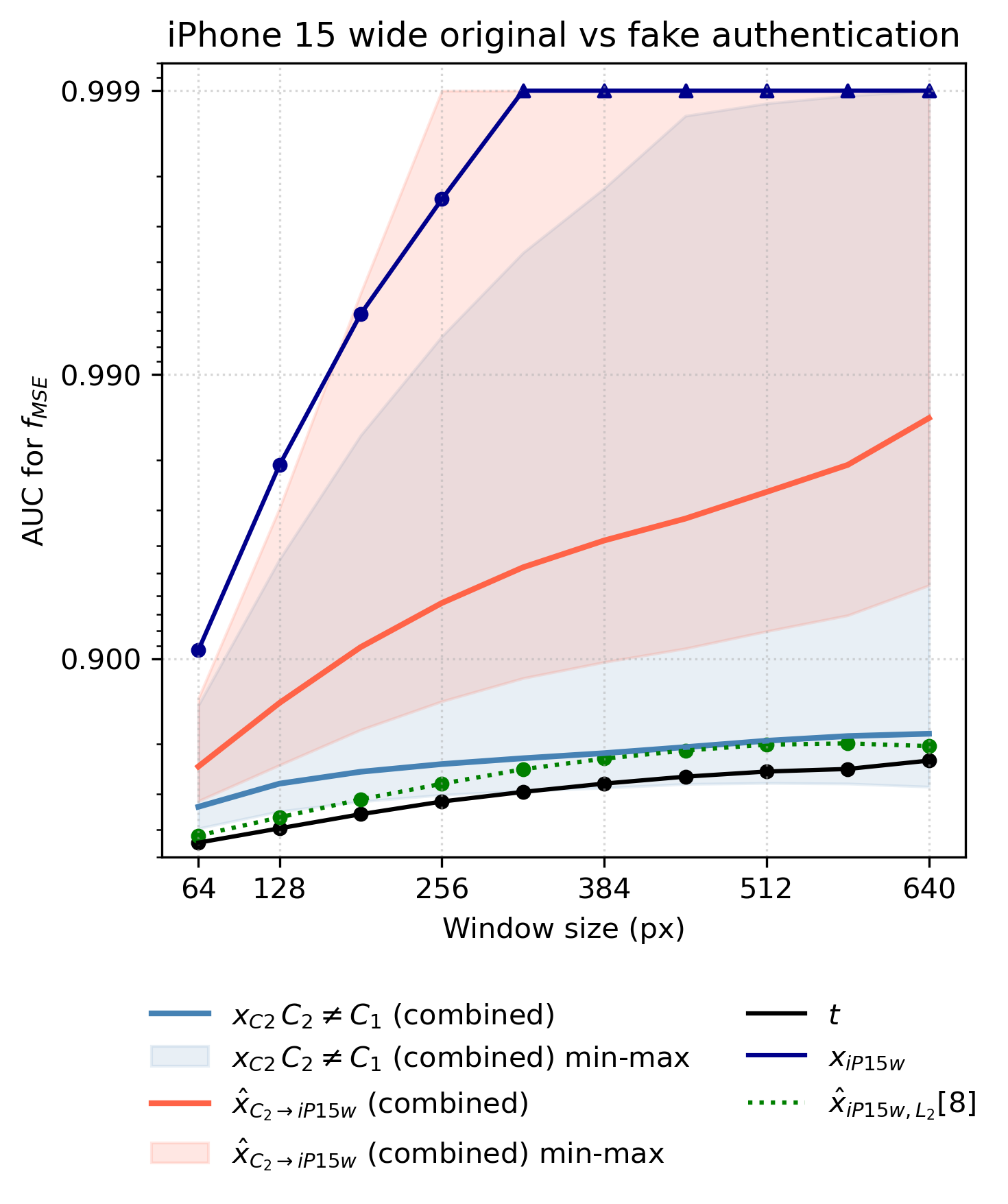}
        \caption{Combined curves}
    \end{subfigure}
\end{figure*}

\begin{figure*}[t]
    \centering
    \begin{subfigure}{.49\textwidth}
        \includegraphics[width=\textwidth]{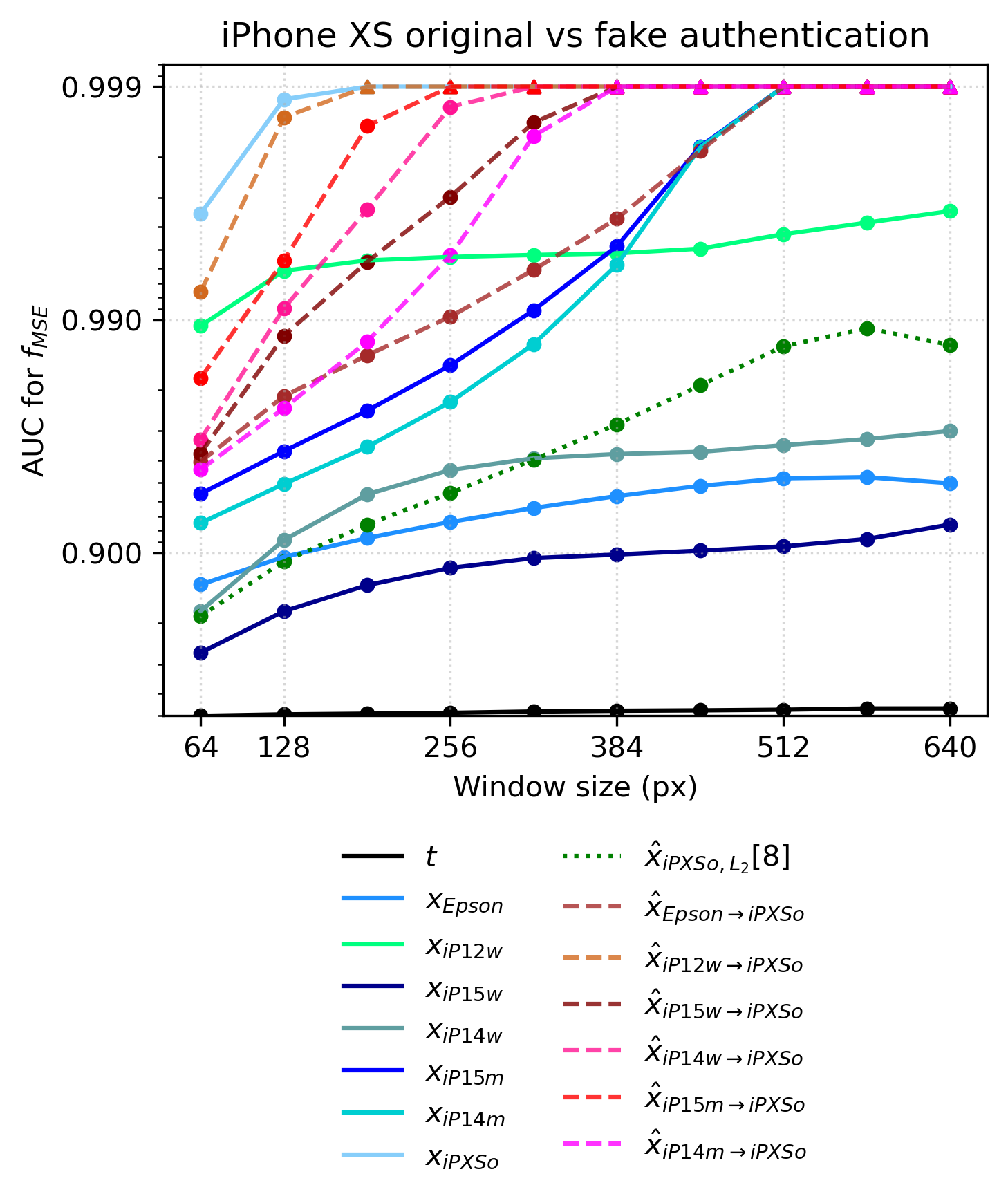}
        \caption{Separate curves}
    \end{subfigure}
    \begin{subfigure}{.49\textwidth}
        \includegraphics[width=\textwidth]{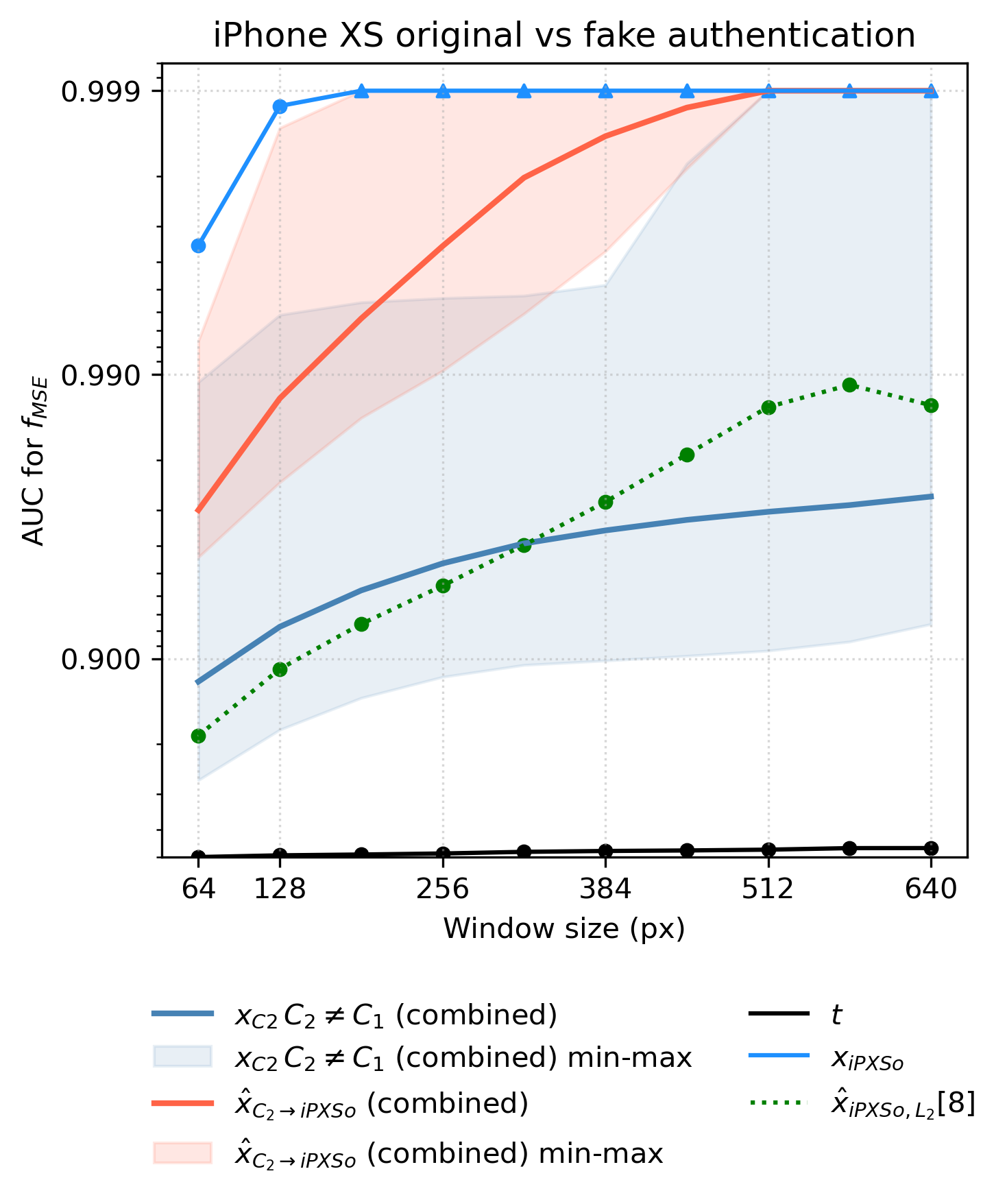}
        \caption{Combined curves}
    \end{subfigure}
\end{figure*}


\begin{figure*}[t]
    \centering
    \begin{subfigure}{.49\textwidth}
        \includegraphics[width=\textwidth]{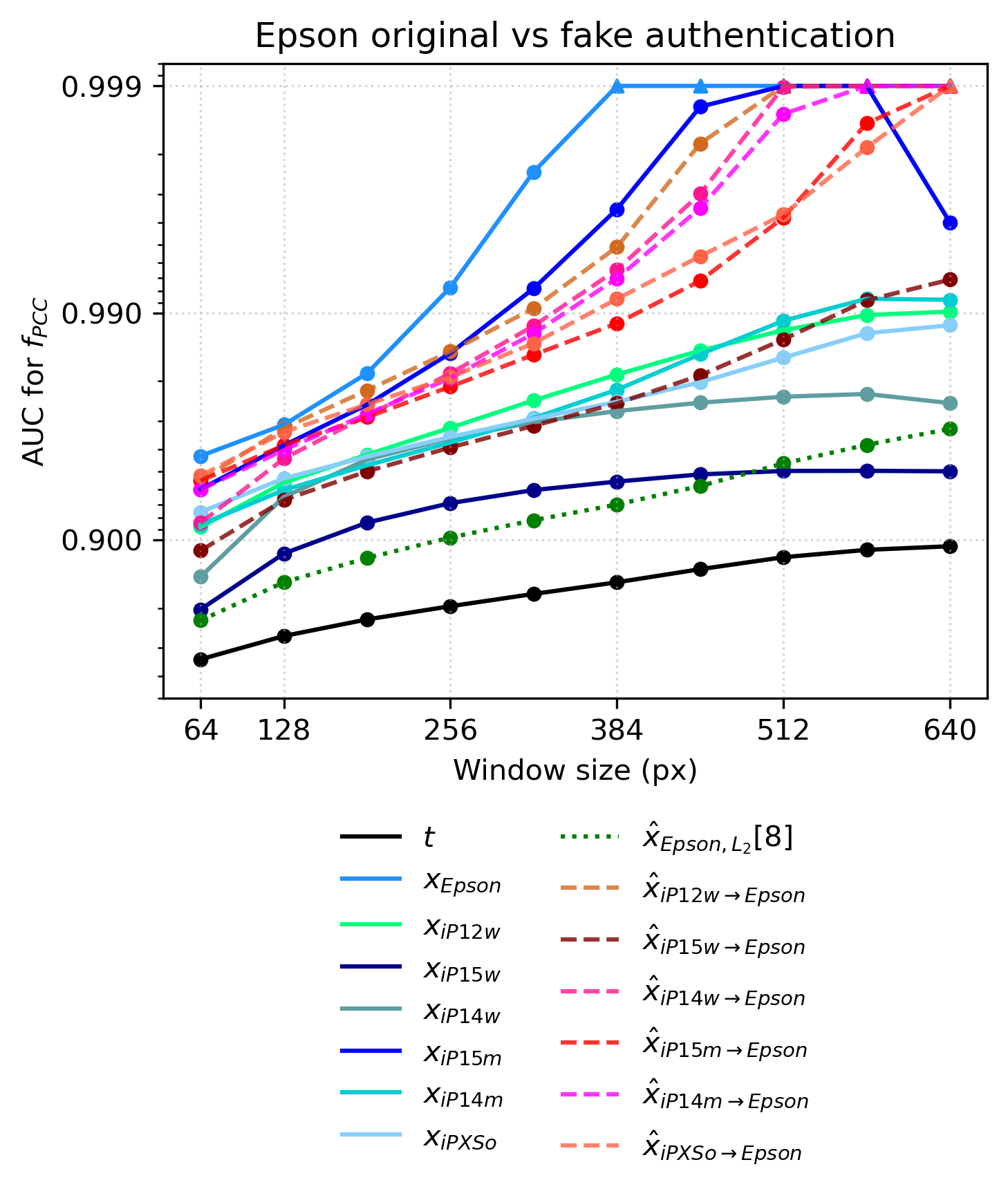}
        \caption{Separate curves}
    \end{subfigure}
    \begin{subfigure}{.49\textwidth}
        \includegraphics[width=\textwidth]{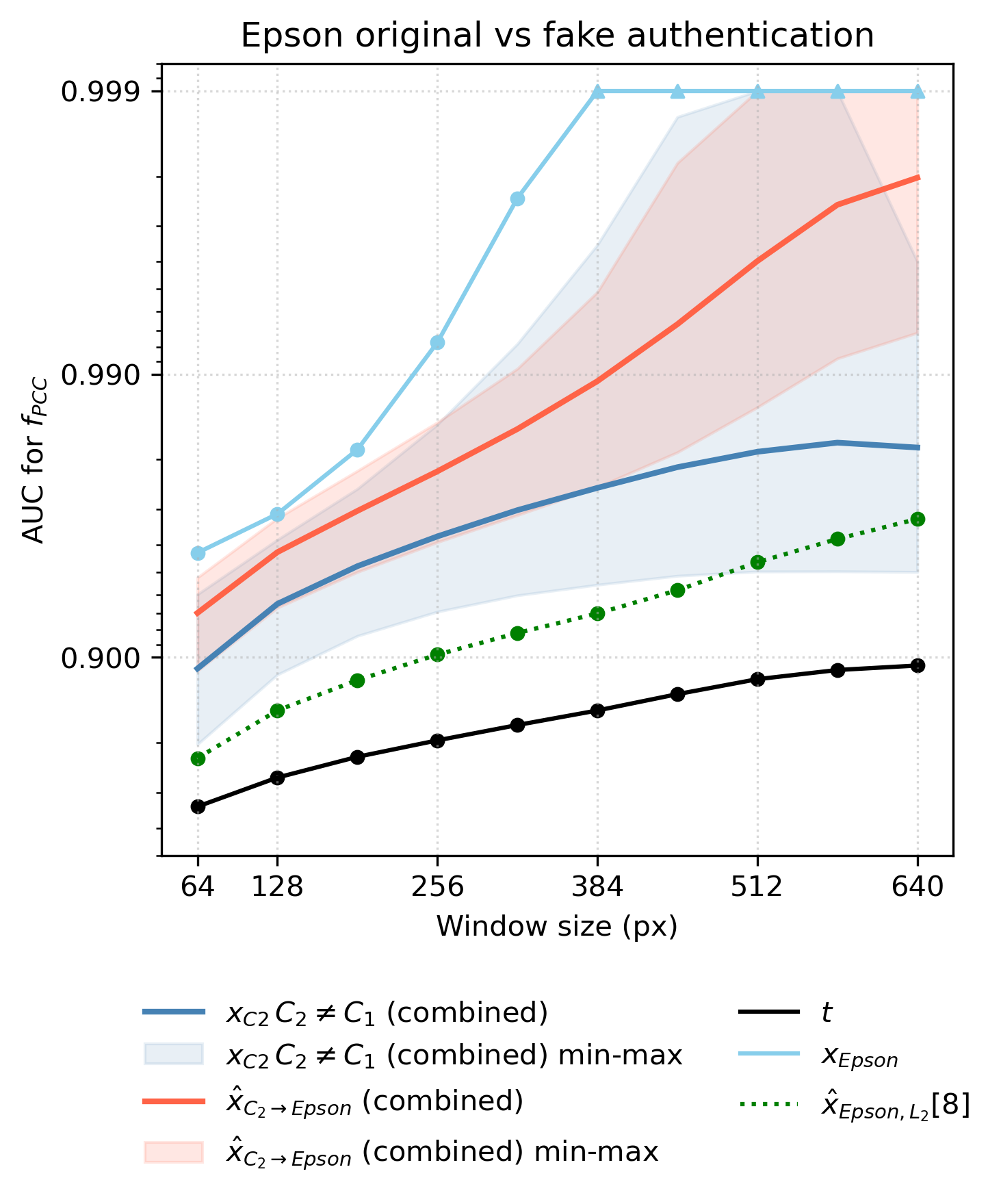}
        \caption{Combined curves}
    \end{subfigure}
\end{figure*}

\begin{figure*}[t]
    \centering
    \begin{subfigure}{.49\textwidth}
        \includegraphics[width=\textwidth]{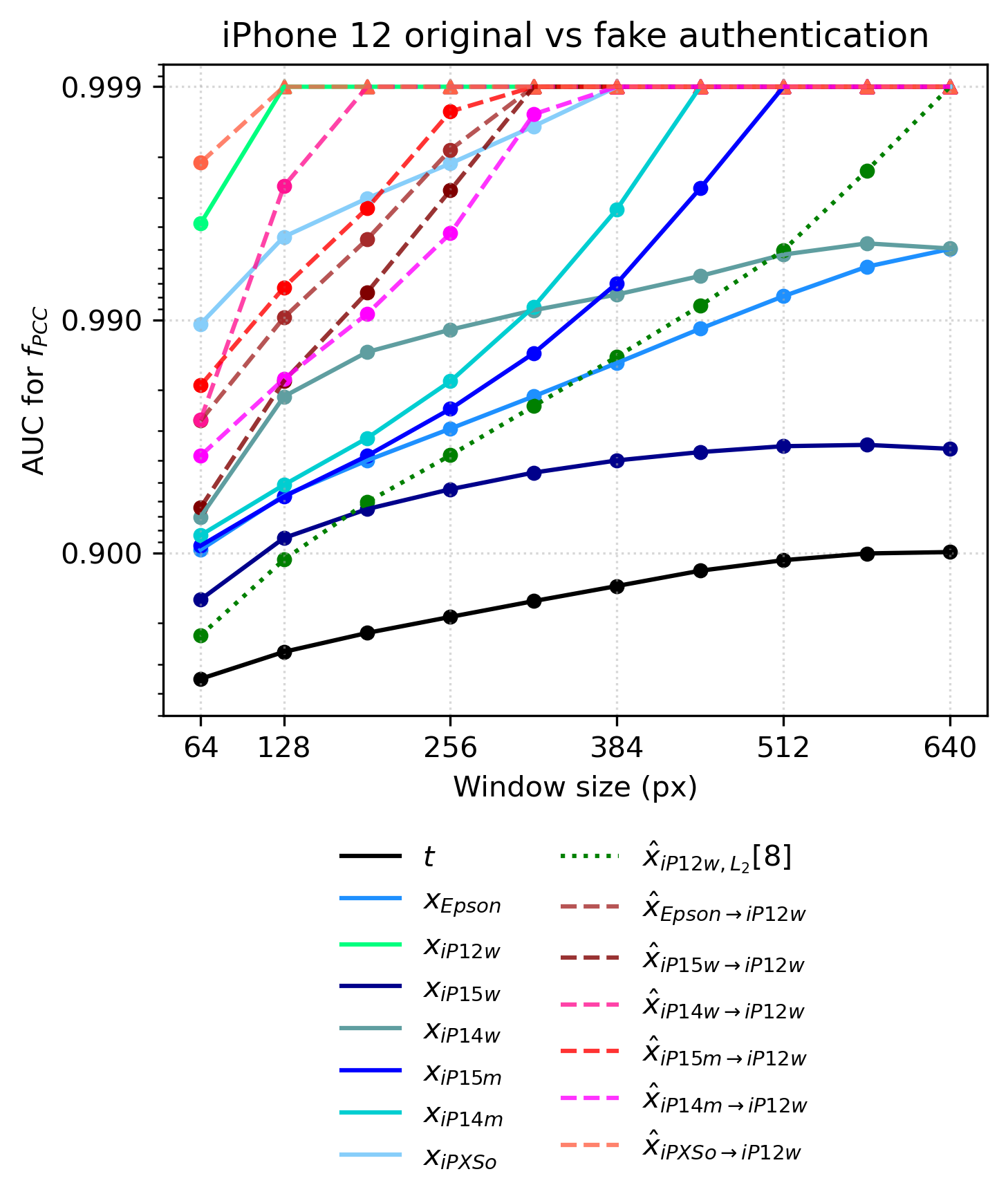}
        \caption{Separate curves}
    \end{subfigure}
    \begin{subfigure}{.49\textwidth}
        \includegraphics[width=\textwidth]{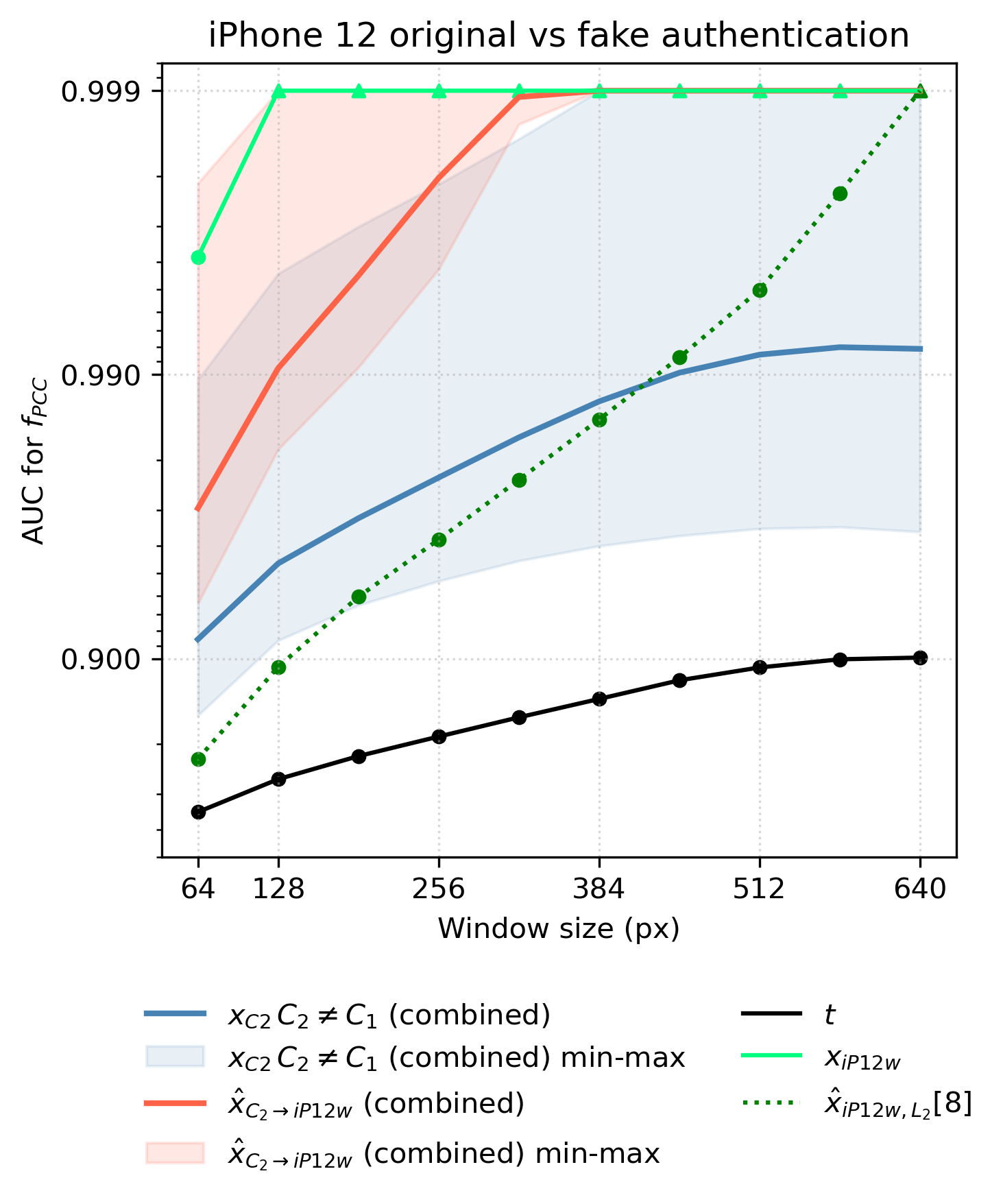}
        \caption{Combined curves}
    \end{subfigure}
\end{figure*}

\begin{figure*}[t]
    \centering
    \begin{subfigure}{.49\textwidth}
        \includegraphics[width=\textwidth]{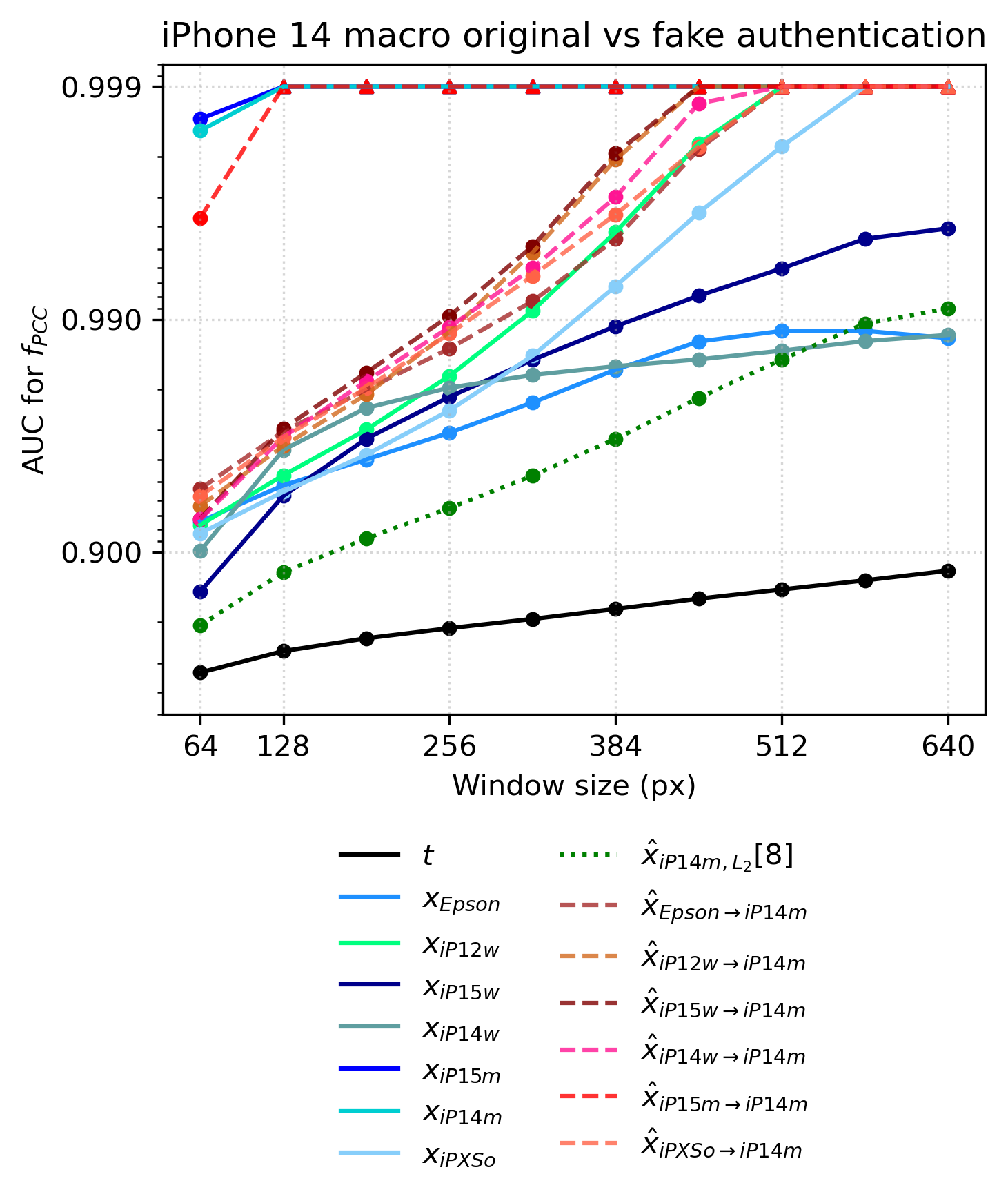}
        \caption{Separate curves}
    \end{subfigure}
    \begin{subfigure}{.49\textwidth}
        \includegraphics[width=\textwidth]{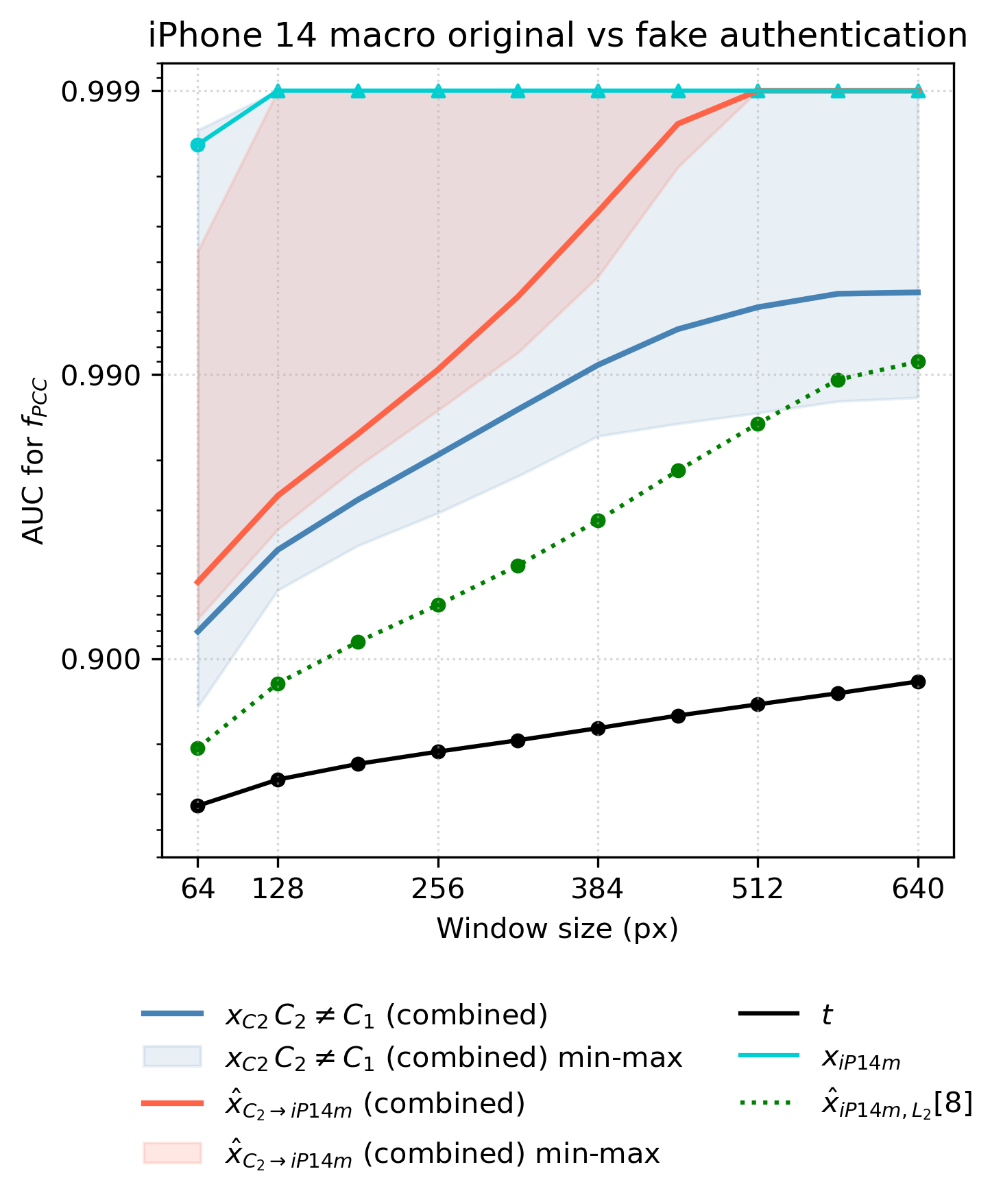}
        \caption{Combined curves}
    \end{subfigure}
\end{figure*}

\begin{figure*}[t]
    \centering
    \begin{subfigure}{.49\textwidth}
        \includegraphics[width=\textwidth]{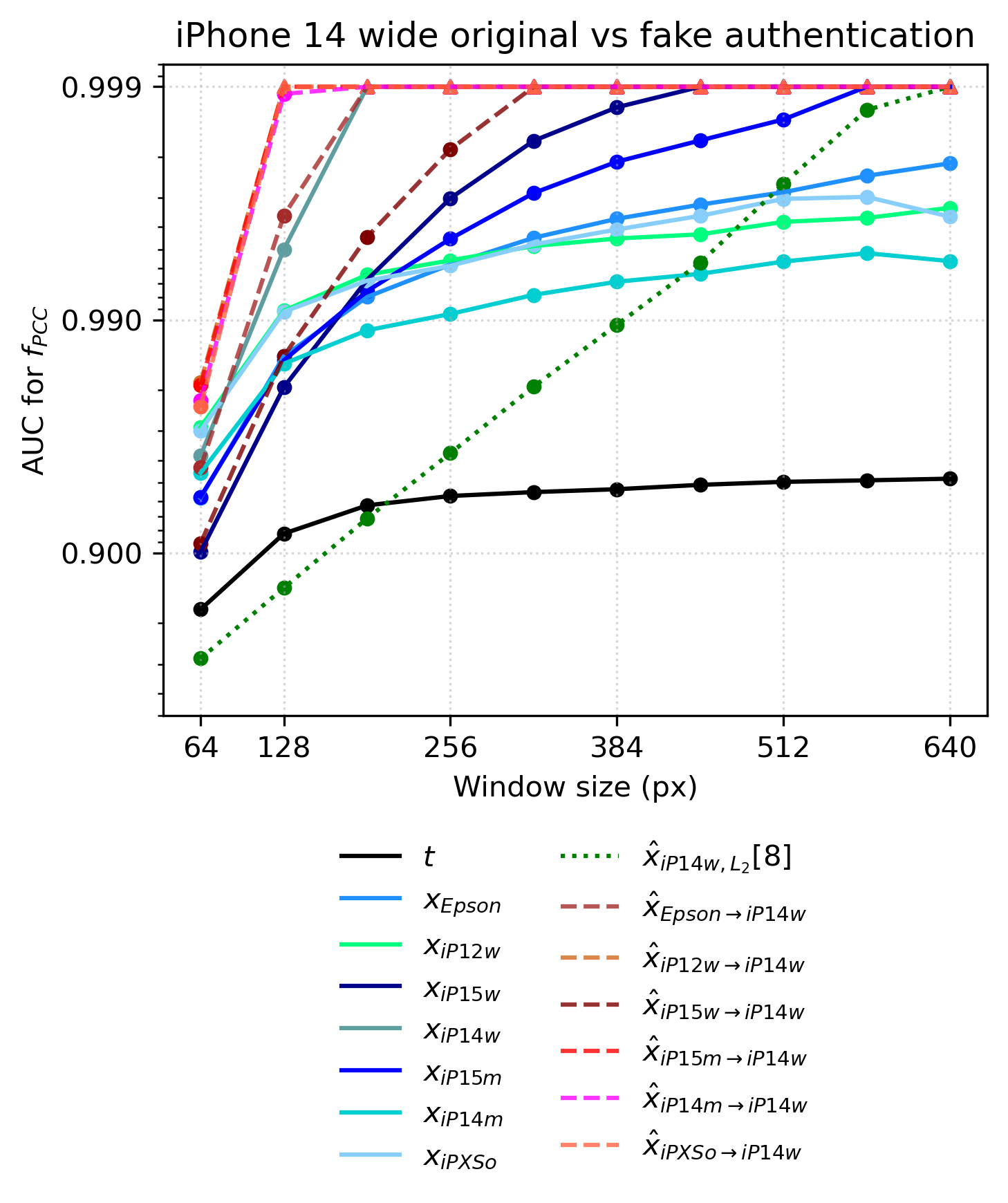}
        \caption{Separate curves}
    \end{subfigure}
    \begin{subfigure}{.49\textwidth}
        \includegraphics[width=\textwidth]{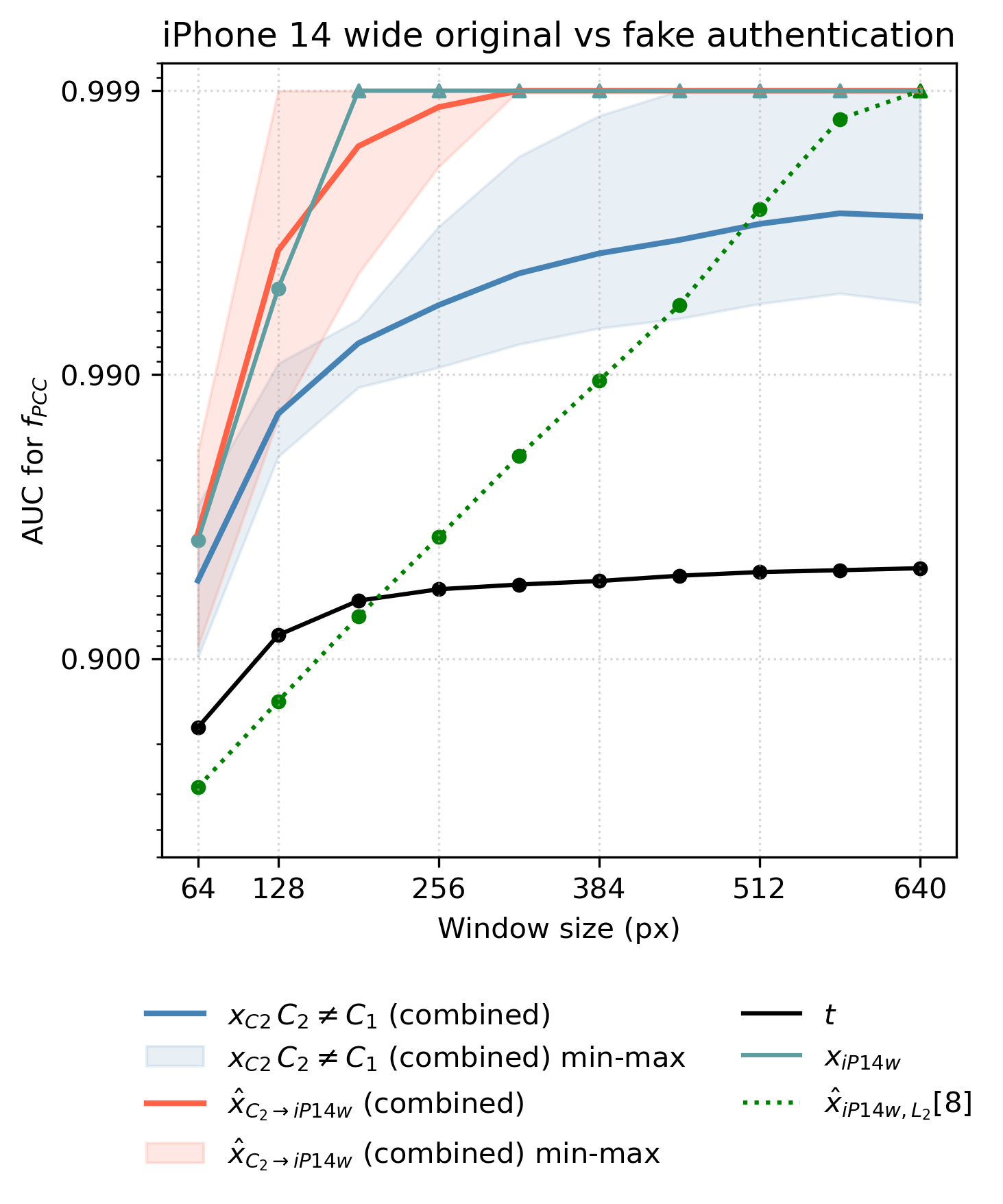}
        \caption{Combined curves}
    \end{subfigure}
\end{figure*}

\begin{figure*}[t]
    \centering
    \begin{subfigure}{.49\textwidth}
        \includegraphics[width=\textwidth]{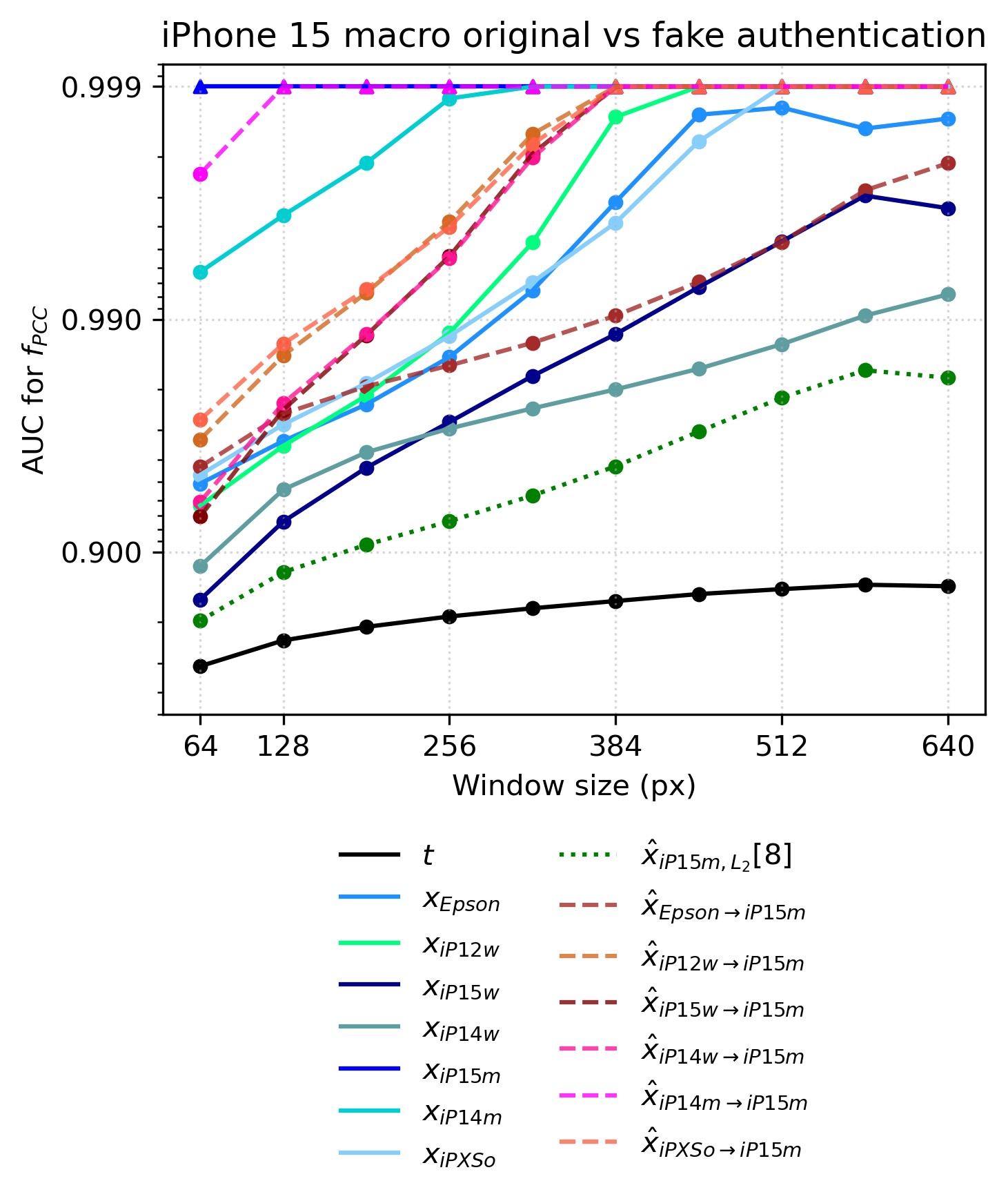}
        \caption{Separate curves}
    \end{subfigure}
    \begin{subfigure}{.49\textwidth}
        \includegraphics[width=\textwidth]{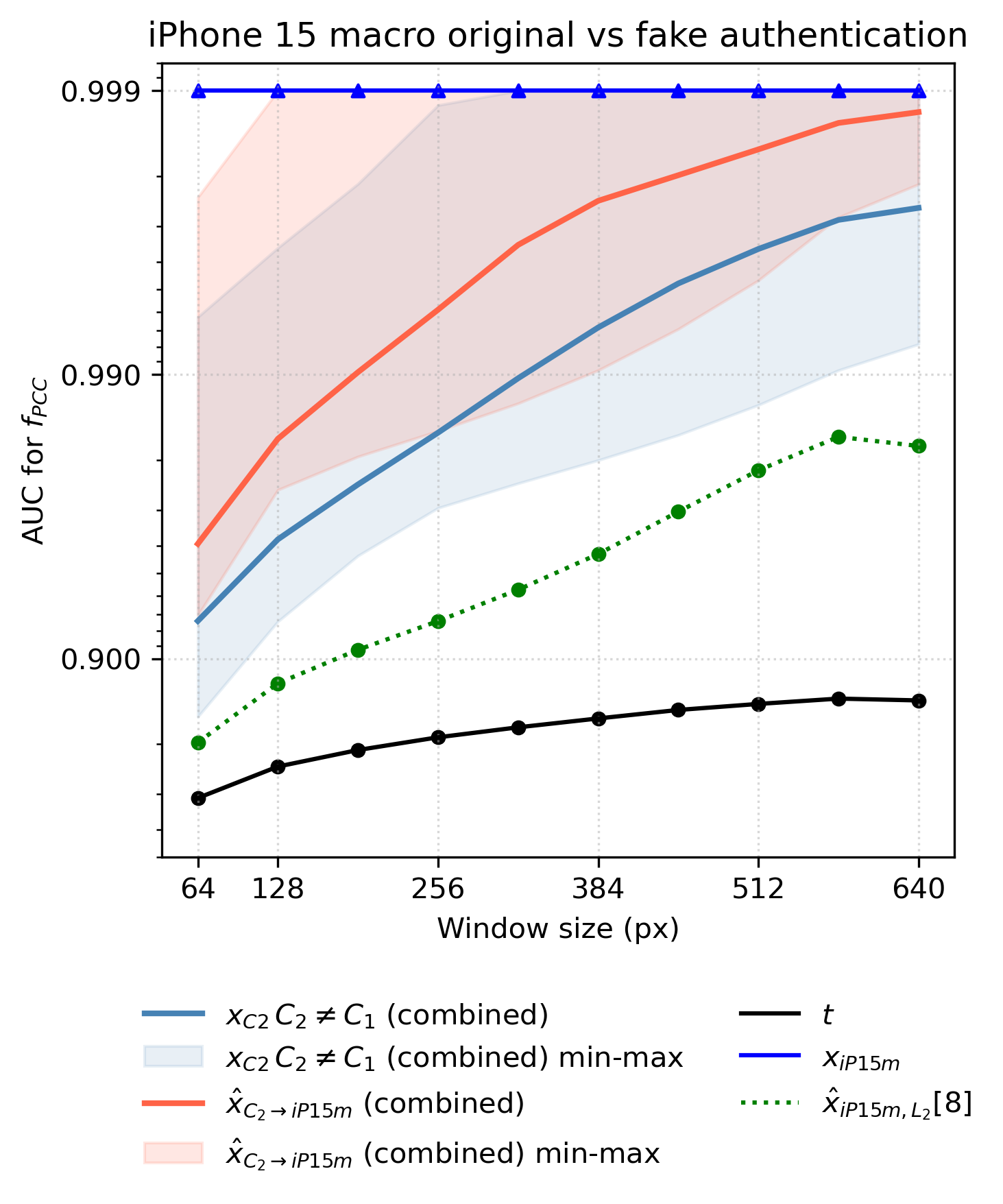}
        \caption{Combined curves}
    \end{subfigure}
\end{figure*}

\begin{figure*}[t]
    \centering
    \begin{subfigure}{.49\textwidth}
        \includegraphics[width=\textwidth]{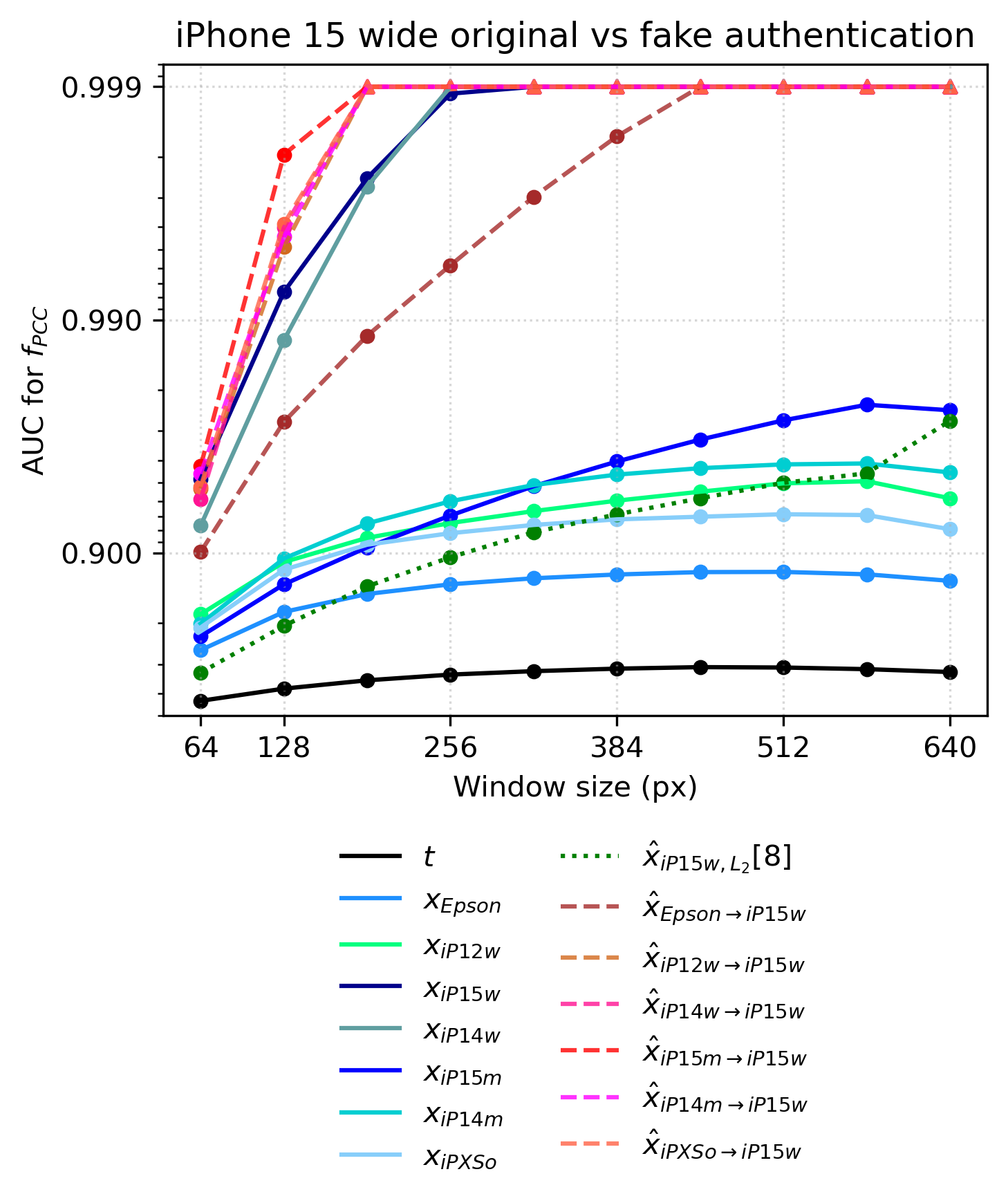}
        \caption{Separate curves}
    \end{subfigure}
    \begin{subfigure}{.49\textwidth}
        \includegraphics[width=\textwidth]{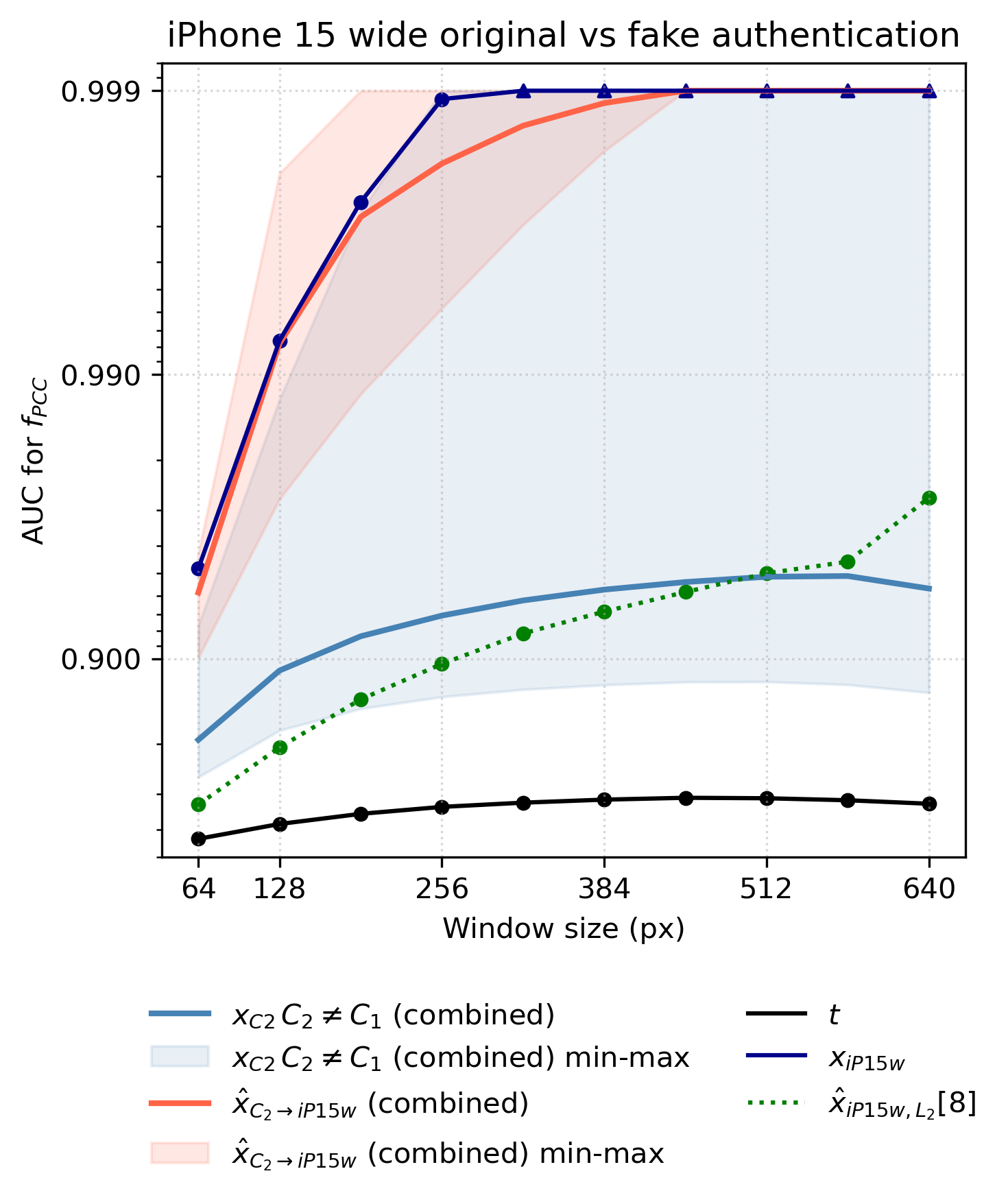}
        \caption{Combined curves}
    \end{subfigure}
\end{figure*}

\begin{figure*}[t]
    \centering
    \begin{subfigure}{.49\textwidth}
        \includegraphics[width=\textwidth]{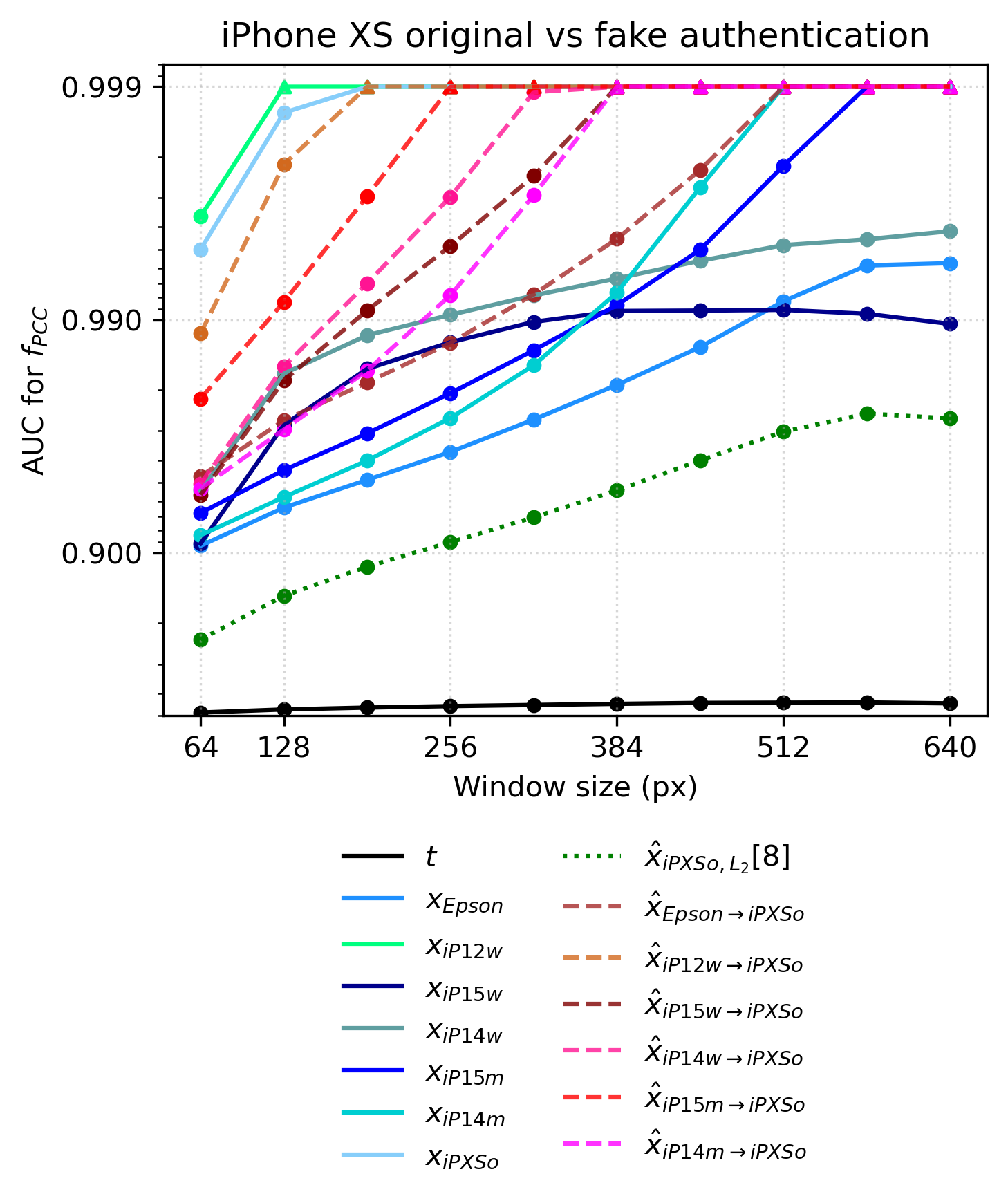}
        \caption{Separate curves}
    \end{subfigure}
    \begin{subfigure}{.49\textwidth}
        \includegraphics[width=\textwidth]{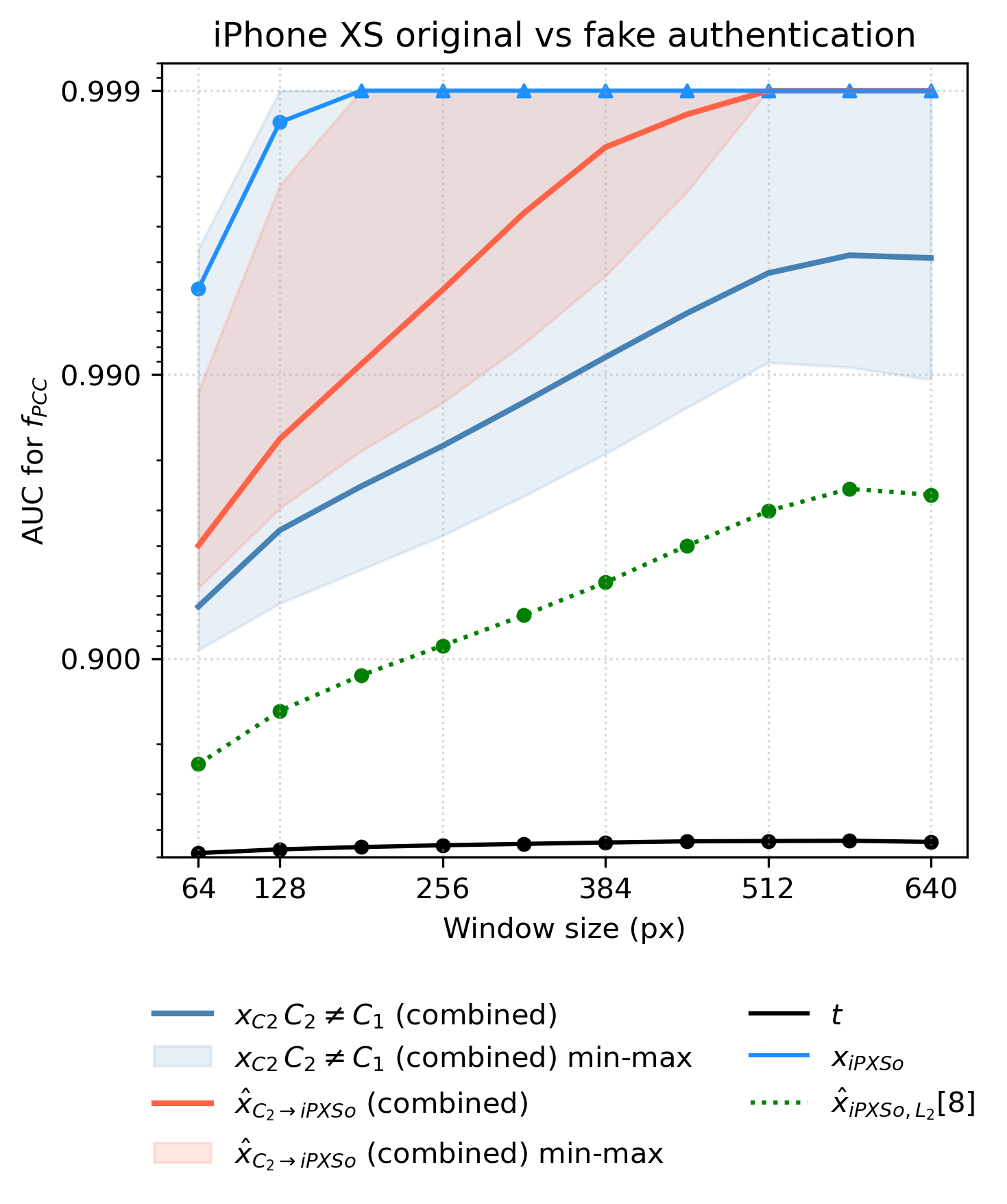}
        \caption{Combined curves}
    \end{subfigure}
    \label{fig:roc}
\end{figure*}


\begin{figure*}[t]
    \centering
    \begin{subfigure}{.49\textwidth}
        \includegraphics[width=\textwidth]{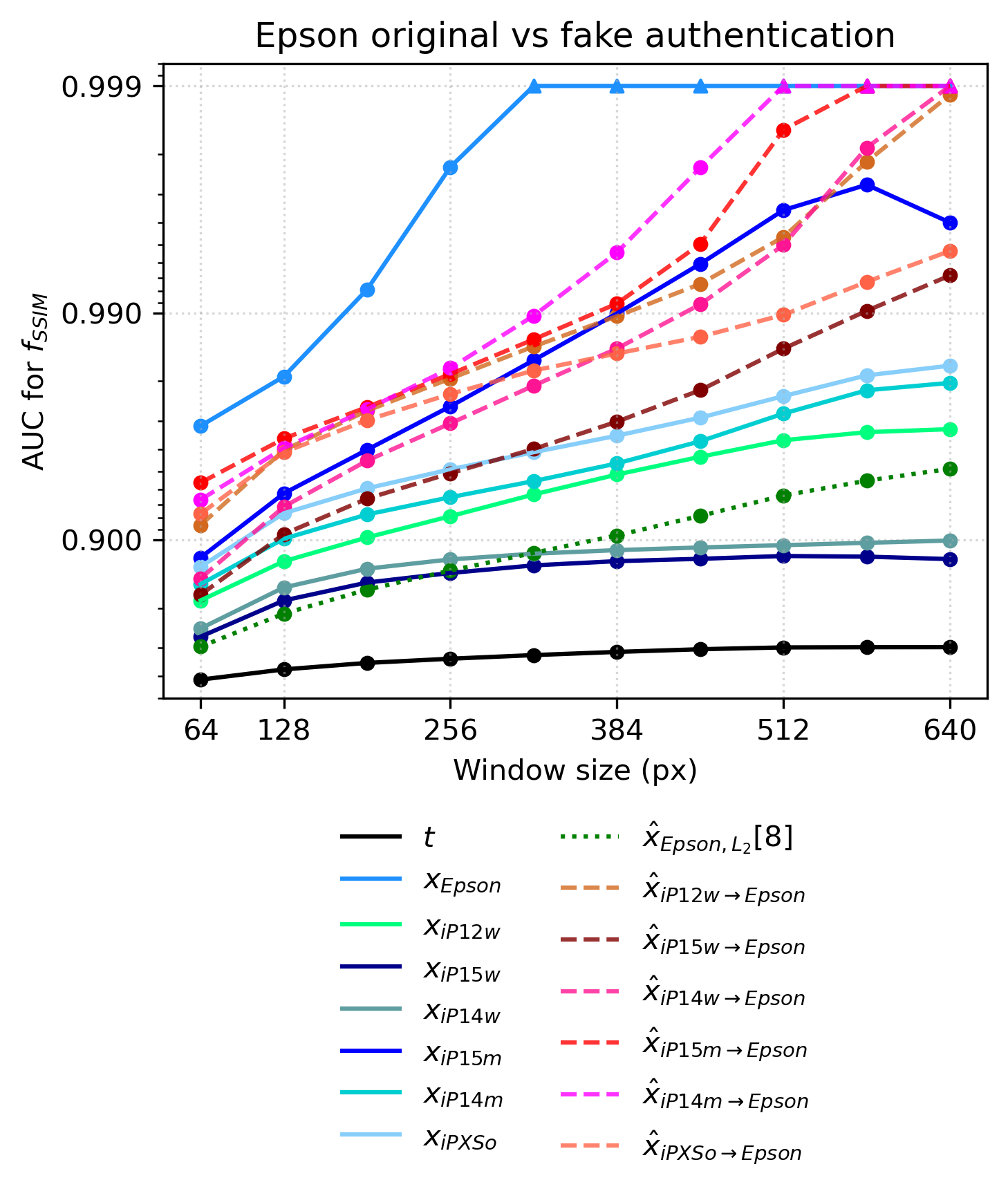}
        \caption{Separate curves}
    \end{subfigure}
    \begin{subfigure}{.49\textwidth}
        \includegraphics[width=\textwidth]{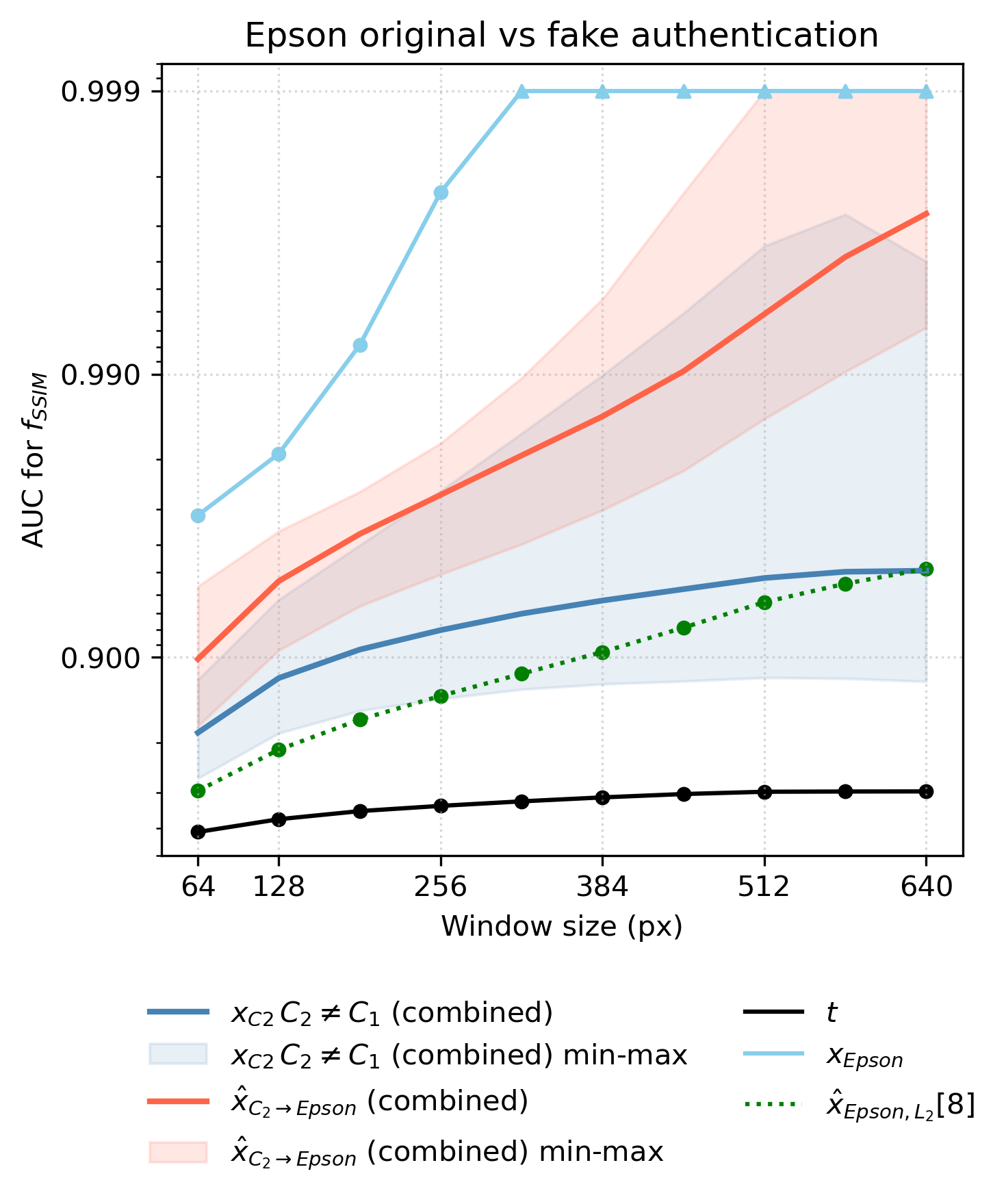}
        \caption{Combined curves}
    \end{subfigure}
\end{figure*}

\begin{figure*}[t]
    \centering
    \begin{subfigure}{.49\textwidth}
        \includegraphics[width=\textwidth]{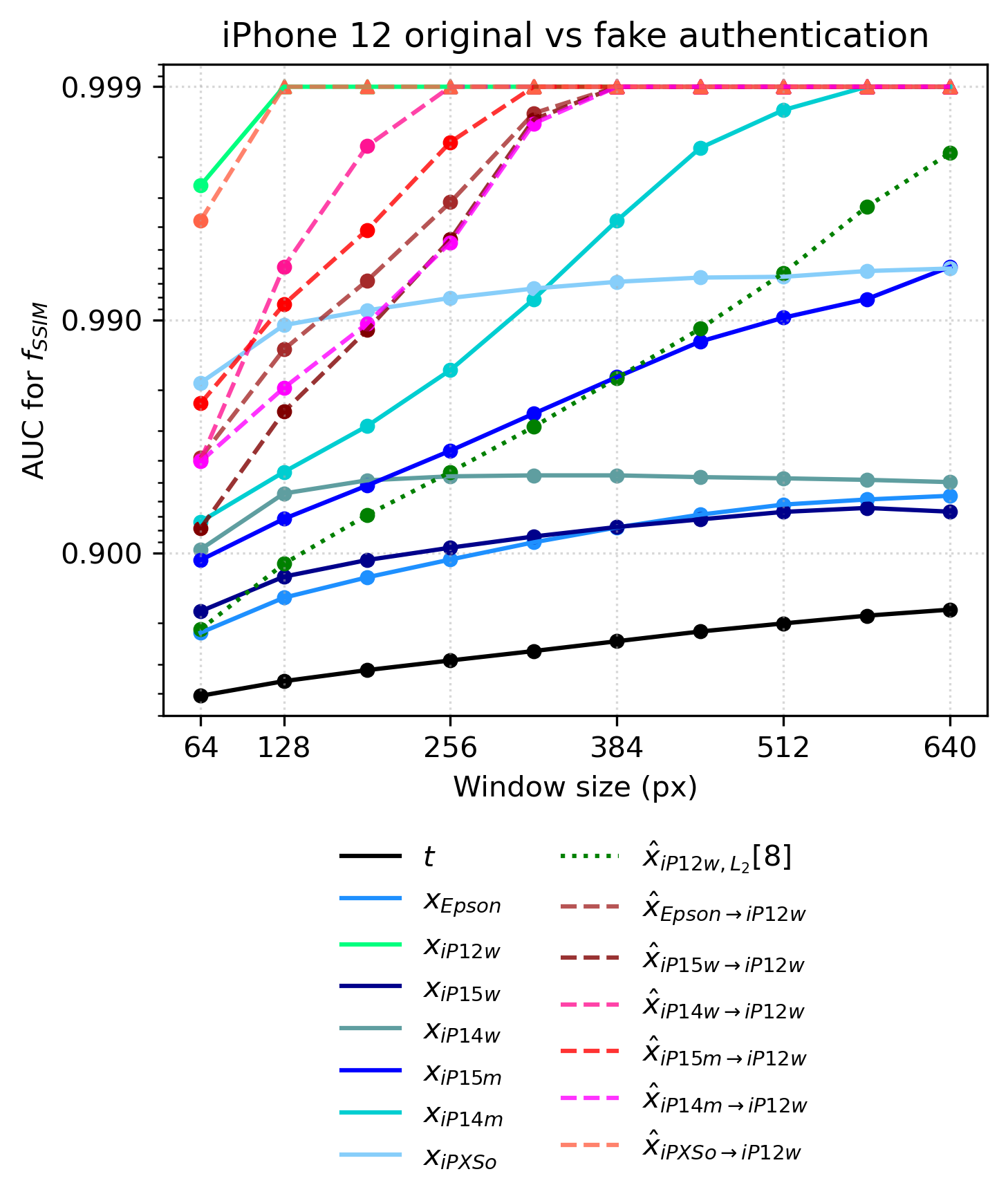}
        \caption{Separate curves}
    \end{subfigure}
    \begin{subfigure}{.49\textwidth}
        \includegraphics[width=\textwidth]{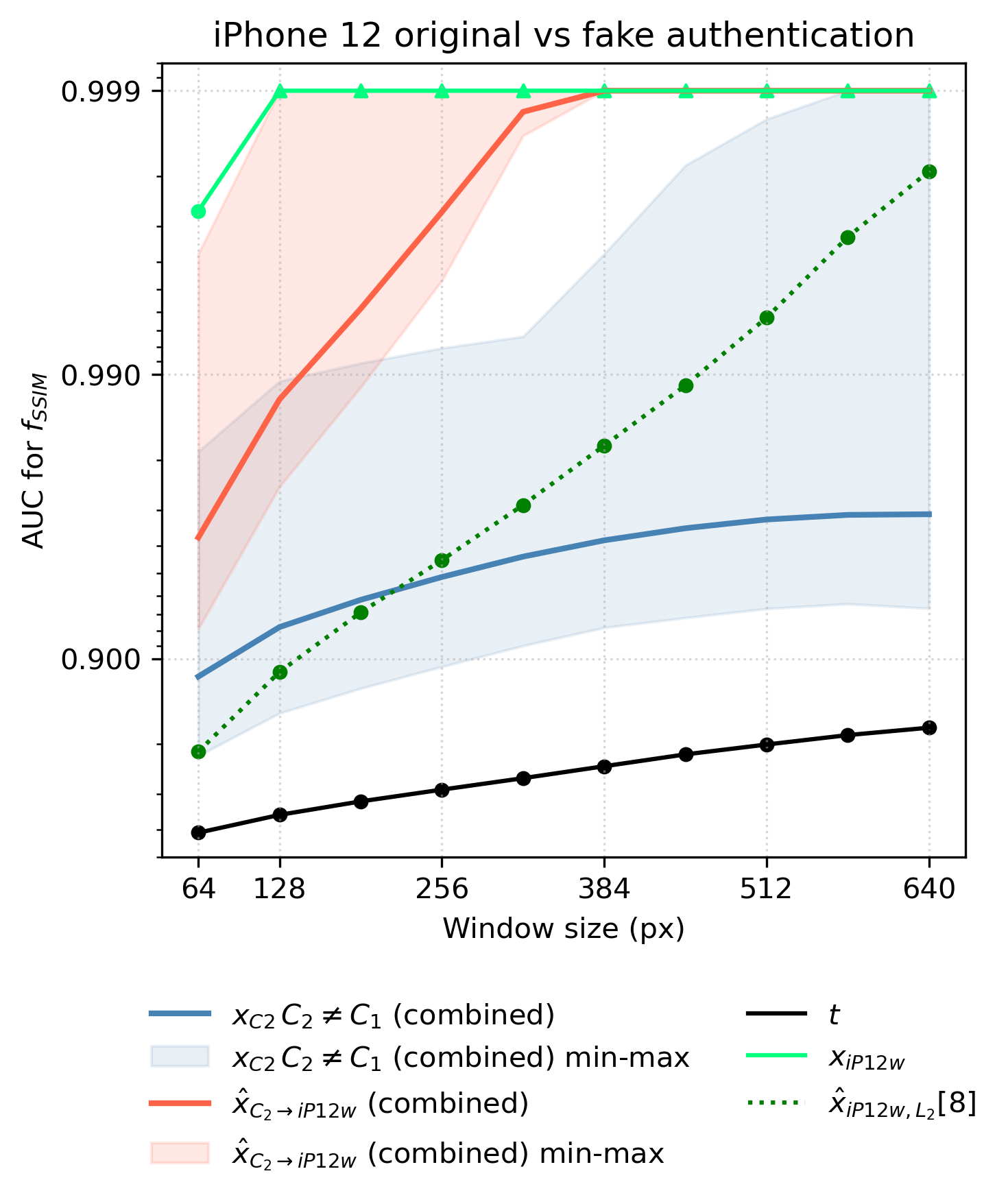}
        \caption{Combined curves}
    \end{subfigure}
\end{figure*}

\begin{figure*}[t]
    \centering
    \begin{subfigure}{.49\textwidth}
        \includegraphics[width=\textwidth]{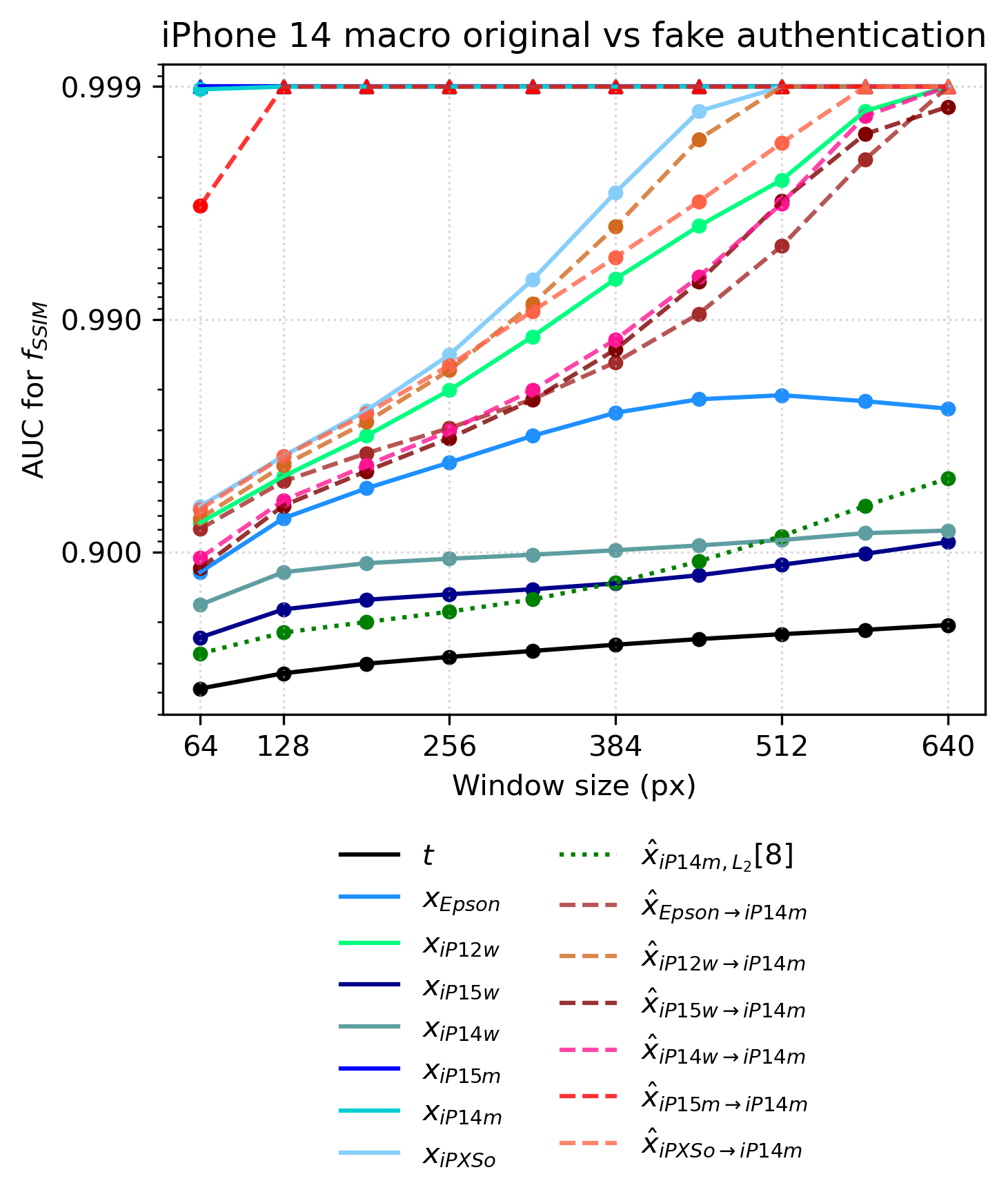}
        \caption{Separate curves}
    \end{subfigure}
    \begin{subfigure}{.49\textwidth}
        \includegraphics[width=\textwidth]{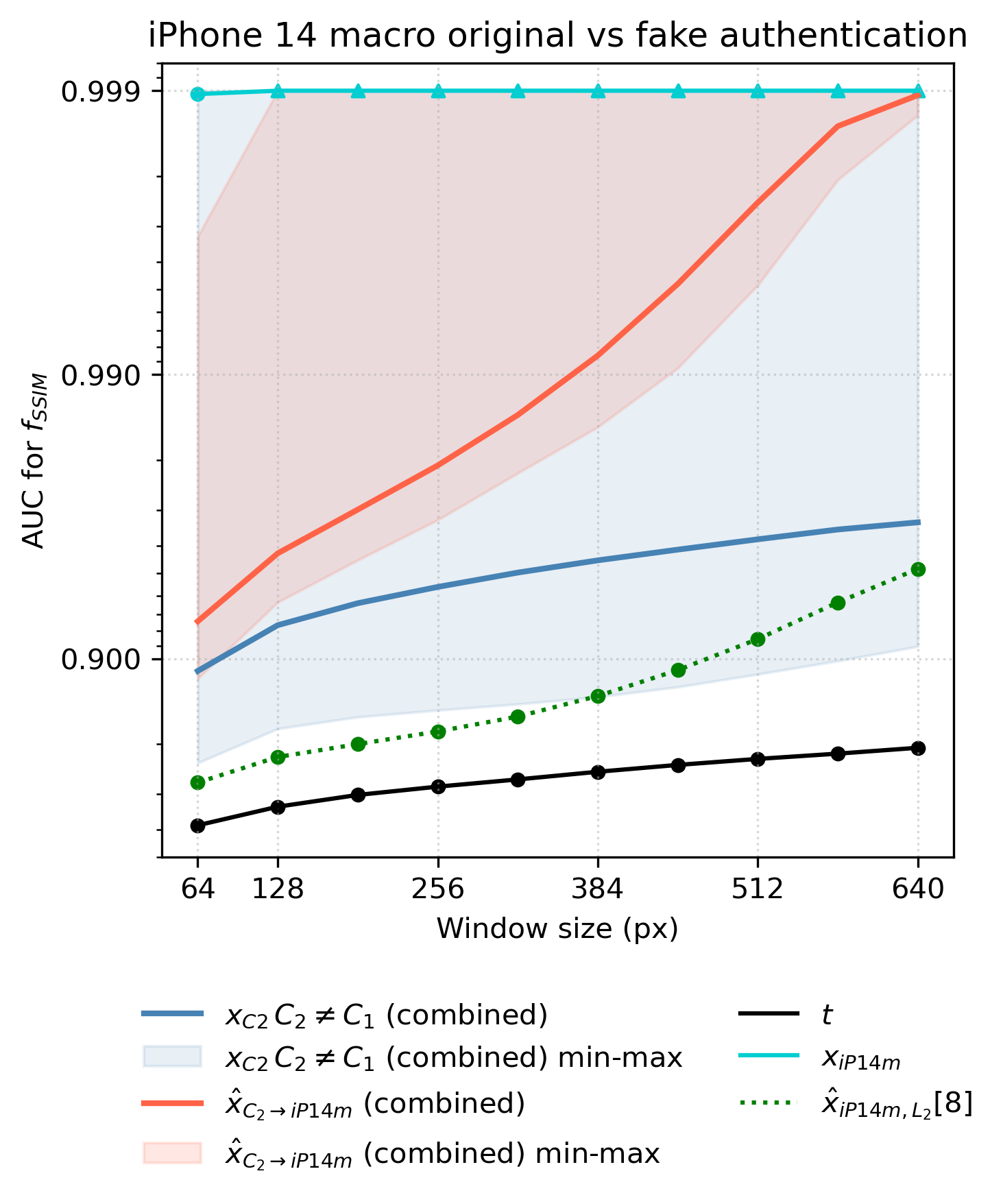}
        \caption{Combined curves}
    \end{subfigure}
\end{figure*}

\begin{figure*}[t]
    \centering
    \begin{subfigure}{.49\textwidth}
        \includegraphics[width=\textwidth]{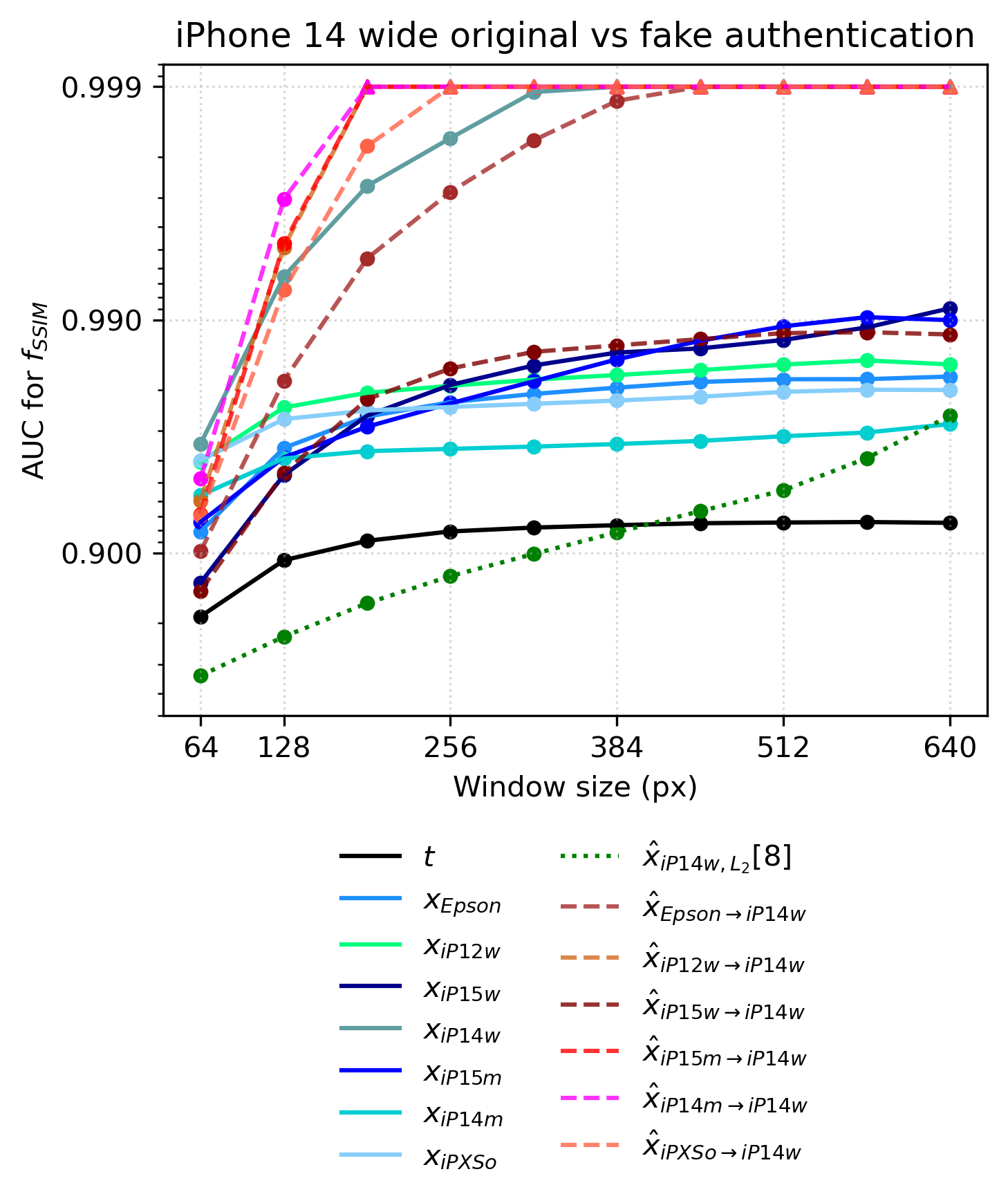}
        \caption{Separate curves}
    \end{subfigure}
    \begin{subfigure}{.49\textwidth}
        \includegraphics[width=\textwidth]{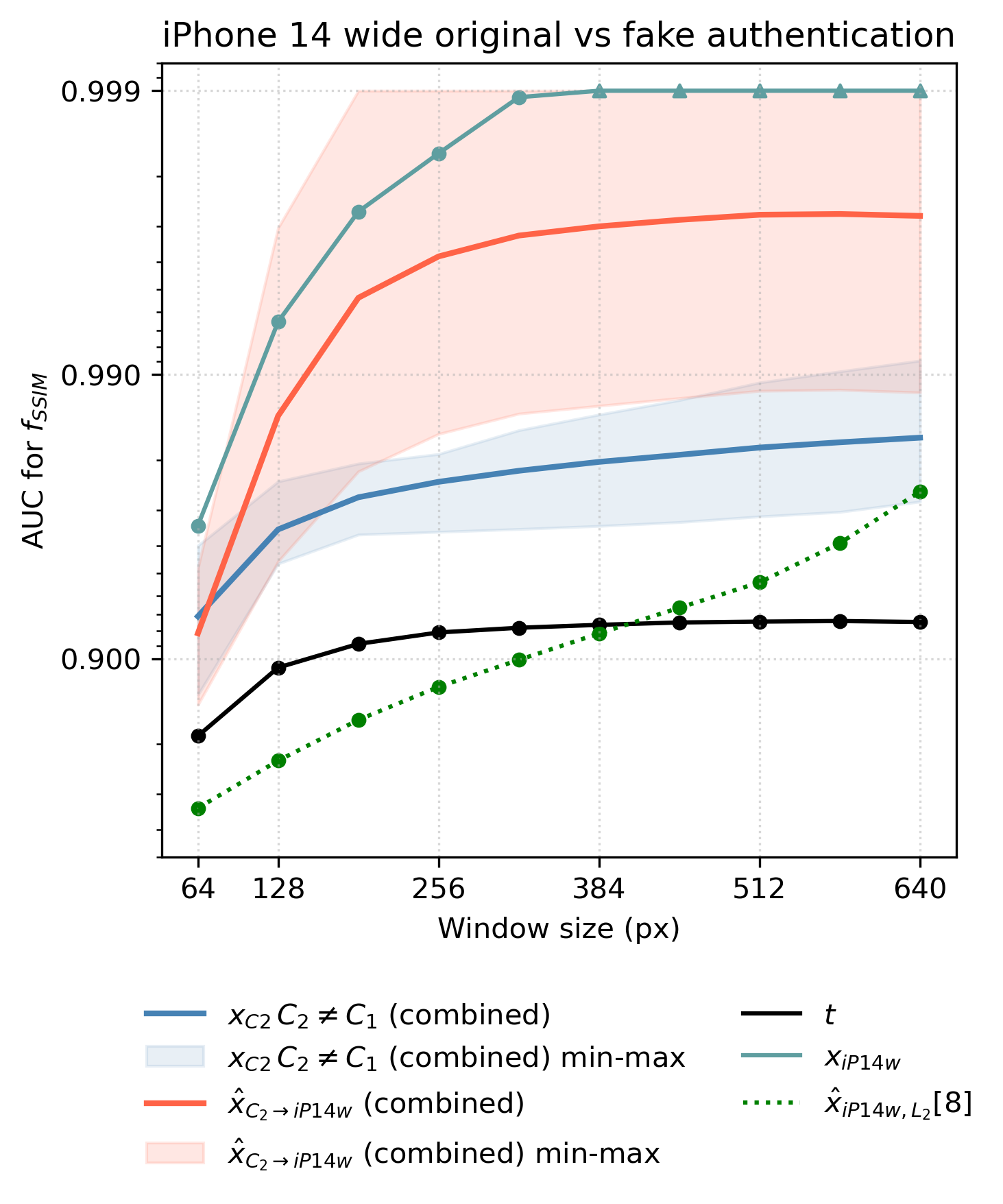}
        \caption{Combined curves}
    \end{subfigure}
\end{figure*}

\begin{figure*}[t]
    \centering
    \begin{subfigure}{.49\textwidth}
        \includegraphics[width=\textwidth]{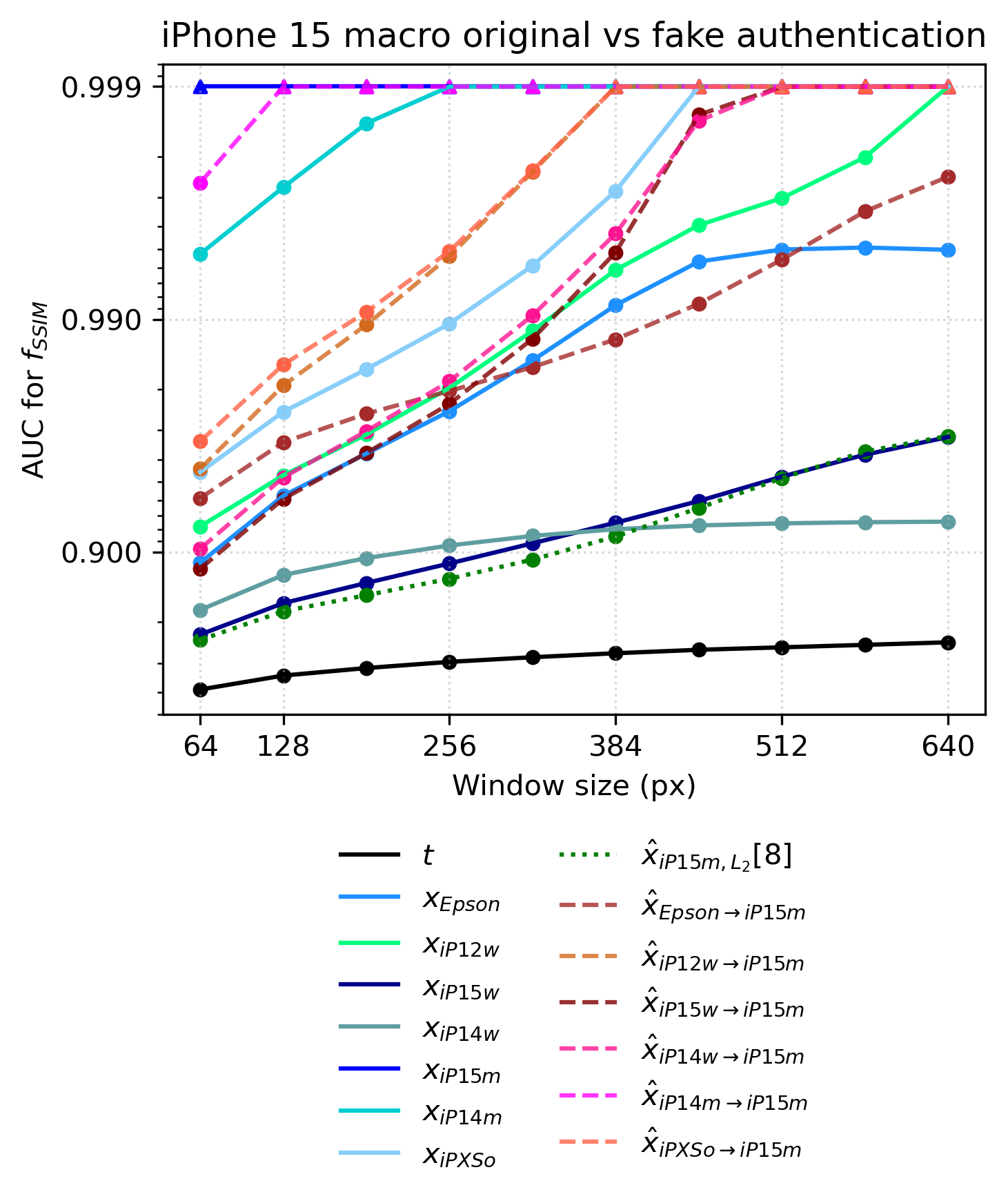}
        \caption{Separate curves}
    \end{subfigure}
    \begin{subfigure}{.49\textwidth}
        \includegraphics[width=\textwidth]{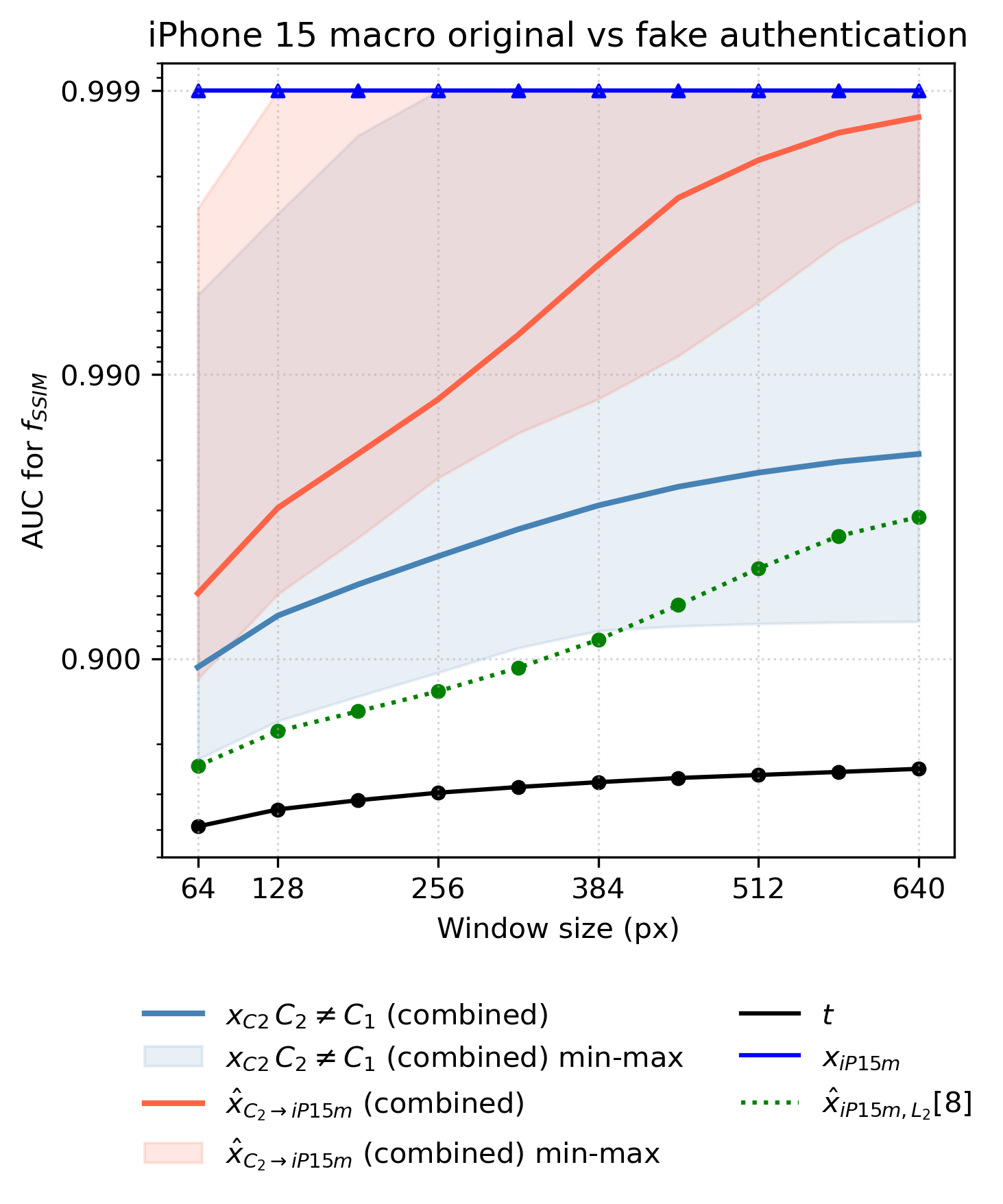}
        \caption{Combined curves}
    \end{subfigure}
\end{figure*}

\begin{figure*}[t]
    \centering
    \begin{subfigure}{.49\textwidth}
        \includegraphics[width=\textwidth]{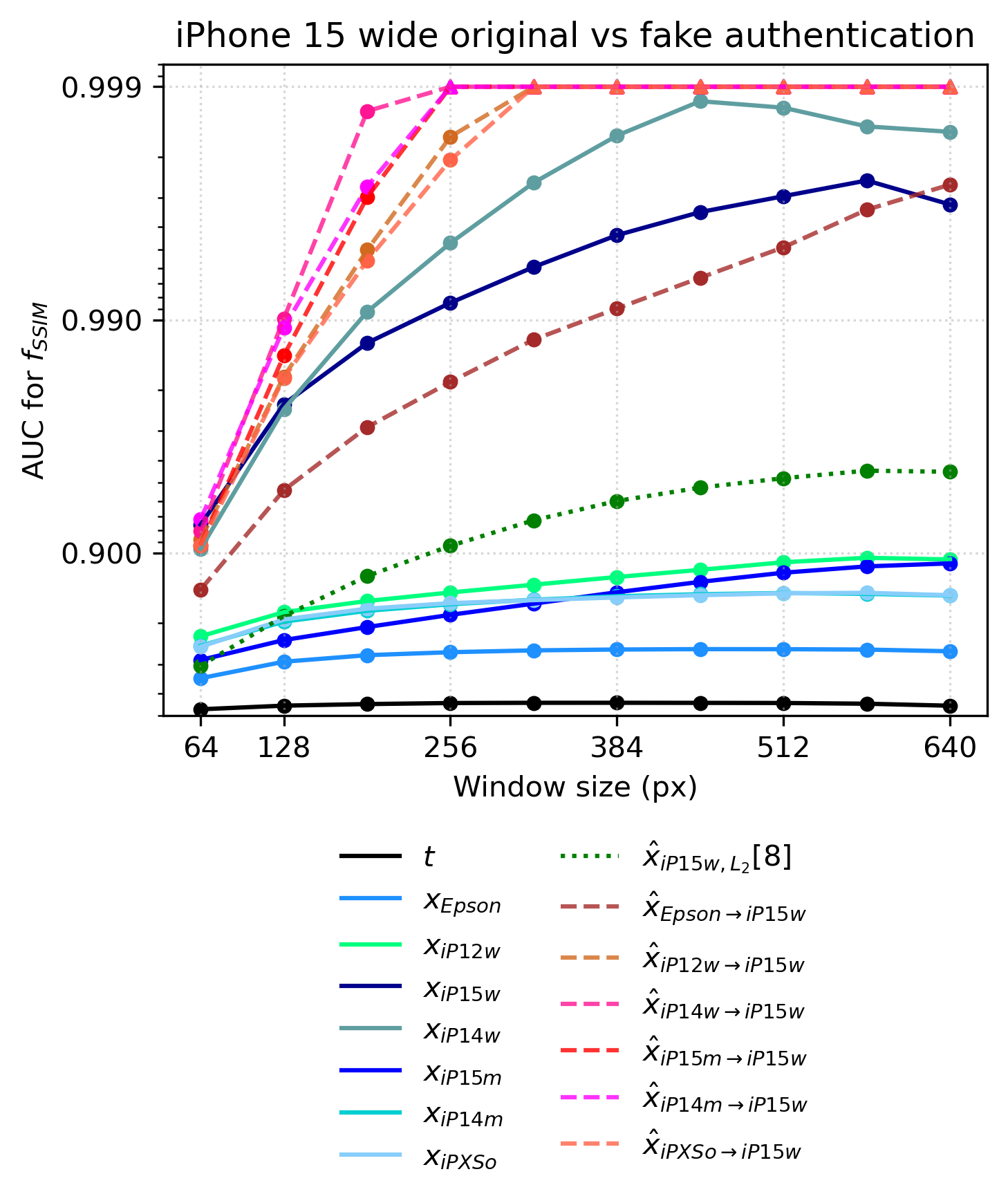}
        \caption{Separate curves}
    \end{subfigure}
    \begin{subfigure}{.49\textwidth}
        \includegraphics[width=\textwidth]{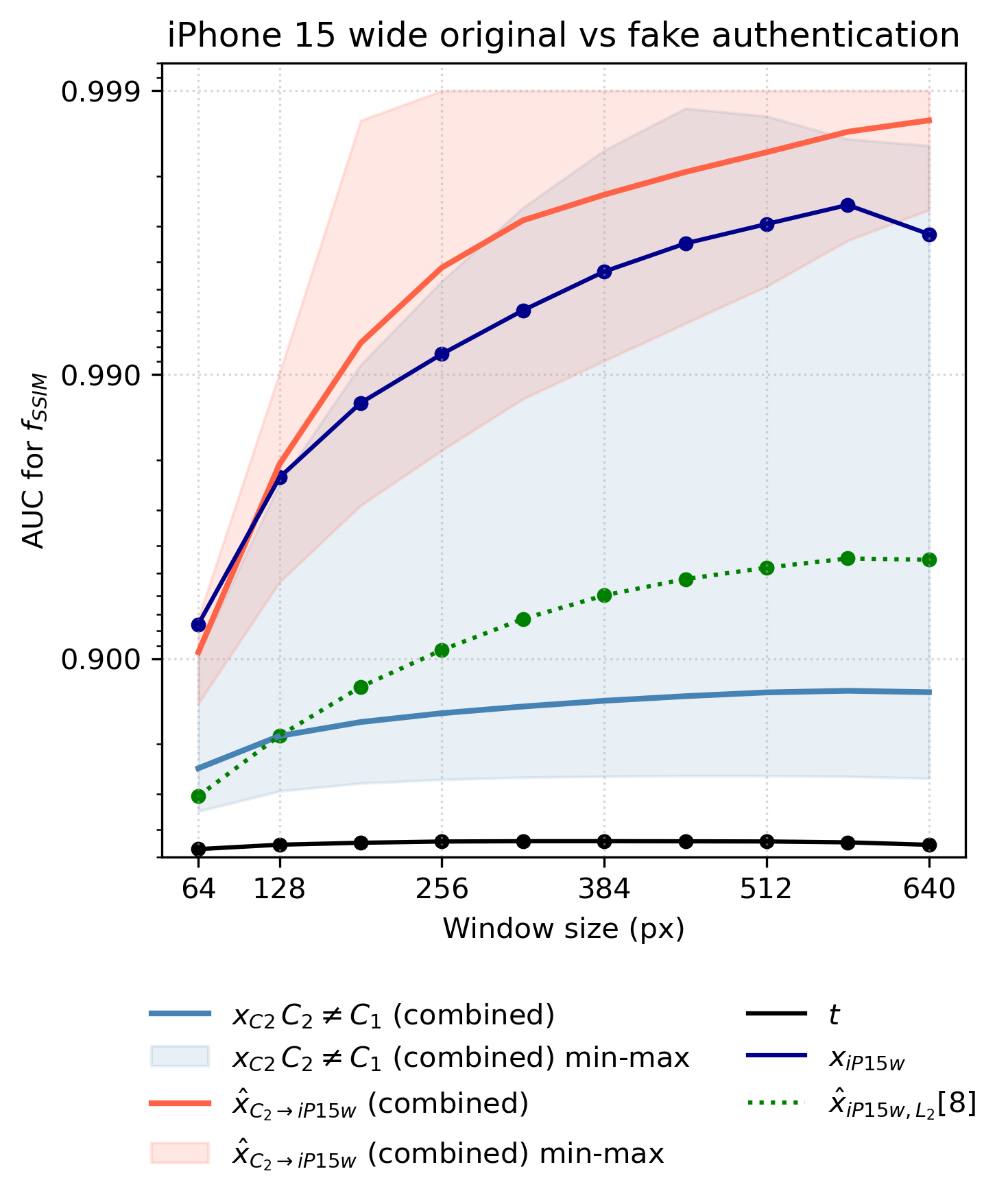}
        \caption{Combined curves}
    \end{subfigure}
\end{figure*}

\begin{figure*}[t]
    \centering
    \begin{subfigure}{.49\textwidth}
        \includegraphics[width=\textwidth]{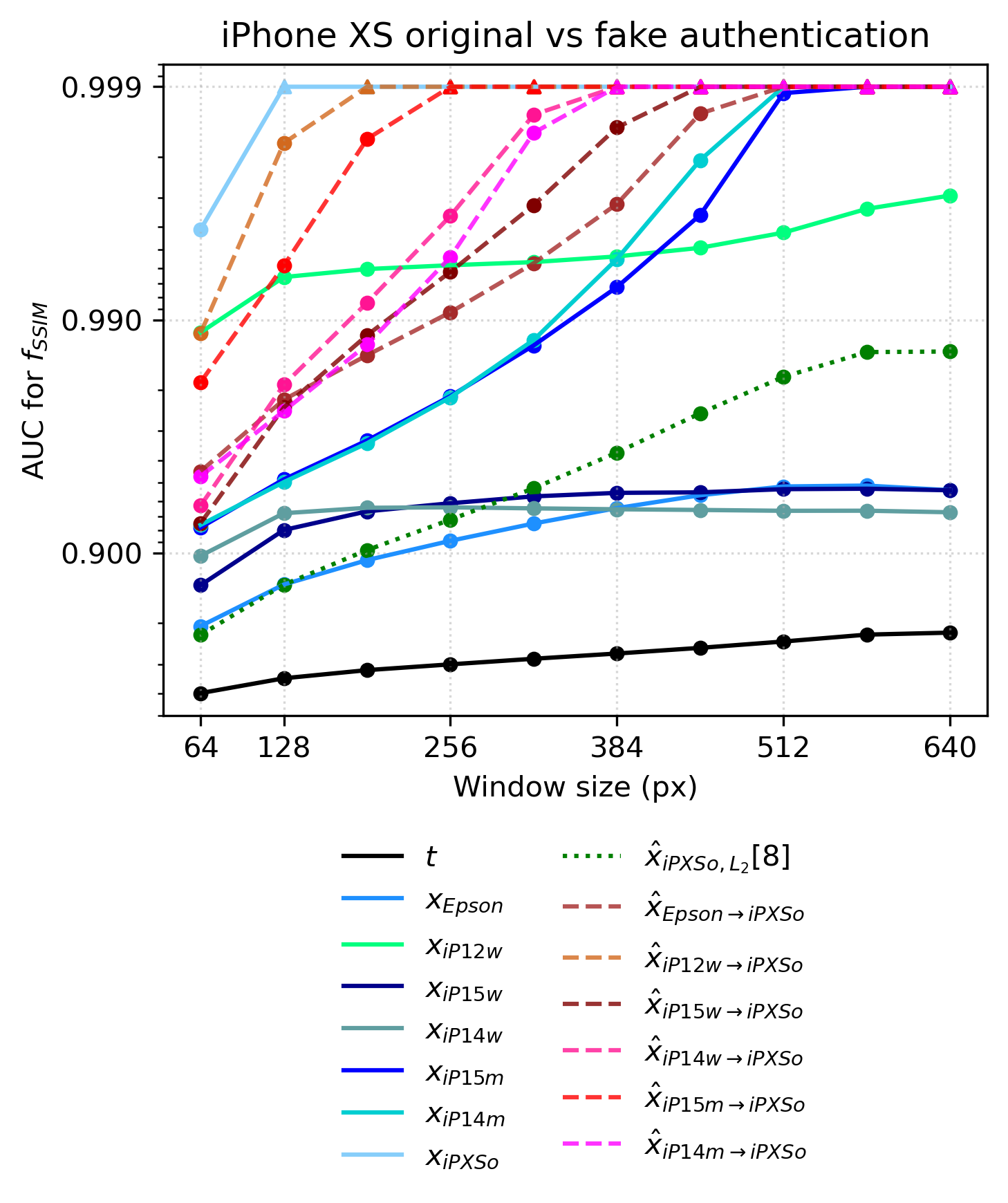}
        \caption{Separate curves}
    \end{subfigure}
    \begin{subfigure}{.49\textwidth}
        \includegraphics[width=\textwidth]{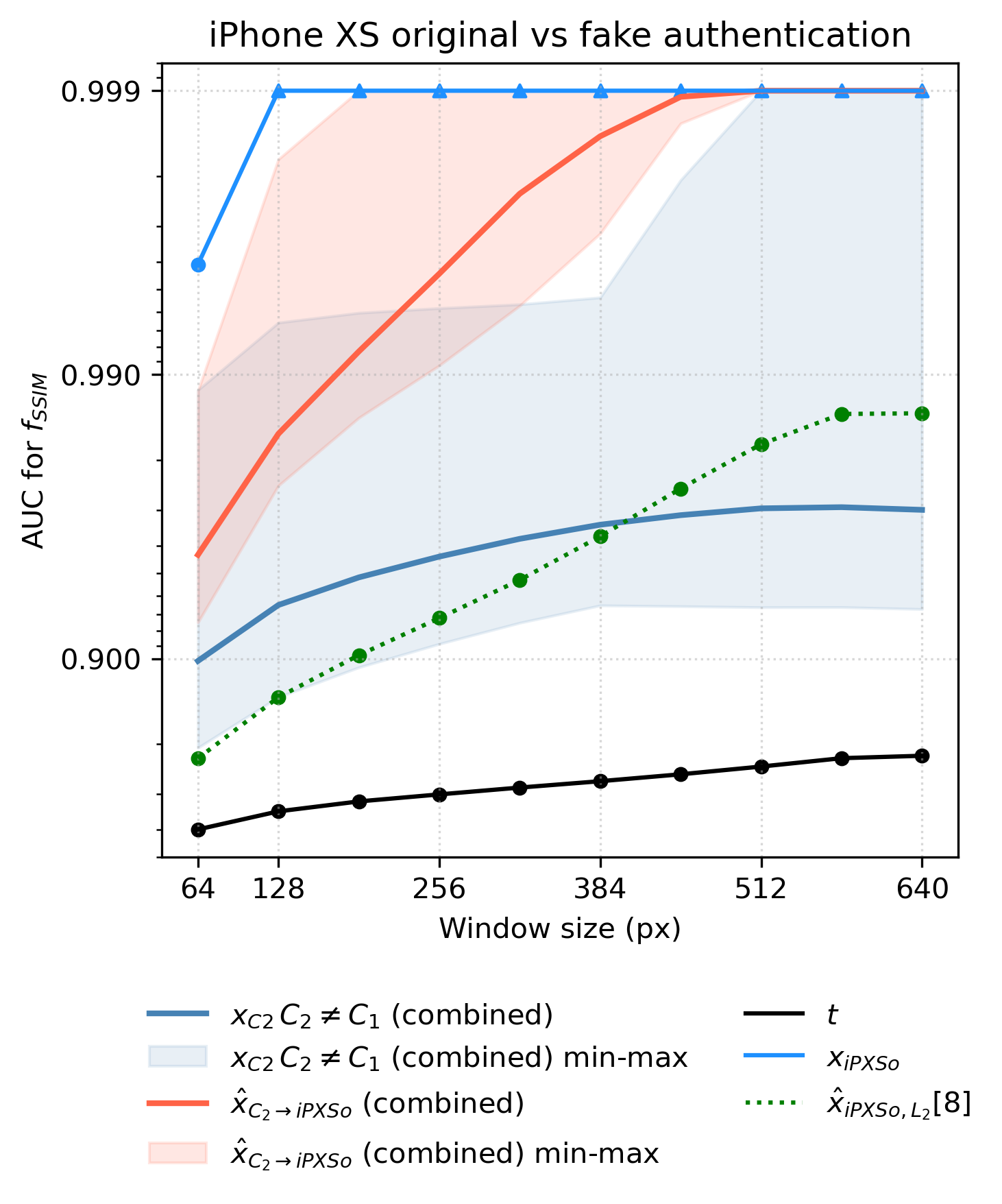}
        \caption{Combined curves}
    \end{subfigure}
\end{figure*}